RESEARCH ARTICLE

# Concept-based Explanations using Non-negative Concept Activation Vectors and Decision Tree for CNN Models

**Running title: Concept-based Decision Tree Explanations**

**Gayda Mutahar[1] | Tim Miller[1]**

[1]School of Computing and Information Systems, The University of Melbourne, Melbourne, Victoria, 3010, Australia

**Correspondence**
Computing and Information Systems, TheUniversity of Melbourne, Melbourne, Victoria, 3010, Australia

Email: gmutahar@student.unimelb.edu.au
Email: gayda.mutahar@gmail.com

This paper evaluates whether training a decision tree based on concepts extracted from a concept-based explainer can increase interpretability for Convolutional Neural Networks (CNNs) models and boost the fidelity and performance of the used explainer. CNNs for computer vision have shown exceptional performance in critical industries. However, it is a significant barrier when deploying CNNs due to their complexity and lack of interpretability. Recent studies to explain computer vision models have shifted from extracting low-level features (pixel-based explanations) to mid-or high-level features (concept-based explanations). The current research direction tends to use extracted features in developing approximation algorithms such as linear or decision tree models to interpret an original model. In this work, we modify one of the state-of-the-art concept-based explanations and propose an alternative framework named TreeICE. We design a systematic evaluation based on the requirements of fidelity (approximate models to original model's labels), performance (approximate models toground-truth labels), and interpretability (meaningful of approximate models to humans). We conduct computational evaluation (for fidelity and performance) and human subject experiments (for interpretability). We find that Tree- ICE outperforms the baseline in interpretability and generates more human-readable explanations in the form of a semantic tree structure. This work features how important to have more understandable explanations when interpretability is crucial.

**KEYWORDS**
Explainable AI, Interpretability, Concept-based explanations, CNNs, Decision tree

## 1 | INTRODUCTION

This paper aims to investigate whether a surrogate concept-based decision tree can improve interpretability, fidelity, and performance requirements. We propose a framework that extracts concepts from a concept-based explainer and facilitates them by a decision tree model to interpret a CNN model. Our method generates both global and local explanations and rethinks one of the recent developed concept-based explanations. We undertake Computational and human subject experiments to validate interpretability, fidelity, and performance, using quantitative and qualitative metrics.

Convolutional Neural Networks (CNNs) are one of the powerful AI algorithms that represent the crux of deep learning algorithms in computer vision [26]. An inherent comprehension of the internal operations is needed in CNNs models, which is why they are described as opaque AI systems. [54]. In contrast, there are a number of machine learning (ML) models are known for their inherent interpretability. These algorithms can be clearly visualized or de-scribed to end-users through plain texts [31]. Rule-based models, such as decision trees, are one of the widely known as interpretable models [7]. Despite the ease of interpret such systems, decision trees have a lower accuracy level than other sophisticated models. As a result, the common trade-off in the field is between achieving high prediction accuracy results while having interpretable and understandable models [20].

Although the complexity of CNNs algorithms plays a key role in improving their prediction performance, it would lead to the black box issue. This problem arises when ML or DL models do not disclose their underlying mechanisms; therefore, trust and transparency concerns appear. To address this phenomenon, we need tools that act as explainable AI models to understand and justify the decisions made by these complex AI systems [35]. Explainable Artificial intelligence (XAI) comes into place as a field of AI that aims to develop methods and techniques to open the black box and promote trust and transparency [34]. XAI seeks to generate understandable, interpretable, and intuitive explanations for AI model decisions while retaining a high level of performance. According to [19], in order to create an interpretable model, it is required to take into consideration a set of desiderata that are interpretability, prediction performance, and fidelity. Interpretability is defined as the understandability degree of a predictive model by humans. While prediction performance or accuracy refers to the model’s ability to predict future data points correctly, fidelity indicates how accurately an interpretable model imitates the behavior of a black-box model.

Many XAI studies currently focus on techniques that retrofit local or approximation algorithms to generate explanations rather than build complex explainable models. Linear and decision tree models are two well-known ap- proximations in the literature [36]. Linear models are intended to measure how each variable is important to the original classifier they are approximating. However, a linear approximation is only helpful if the users are aware of its shortcomings, including situations in which it may fail to explain the original classifier correctly. The two fundamental difficulties linear models suffer from are curvature and dependencies [36]. Linear approximations also suffer from generalization problems as their primary value and reliability for non-expert users are extremely dubious, especially for those who depend on the system's decisions [36]. On top of these issues, numerical weights that indicate feature importance are sometimes difficult to interpret and justify [40]. Thus, alternative approximation methods may be advantageous from a usability and interpretability perspective. Surrogate models such as rule lists, explanatory graphs, or decision trees from DL/ML models has been an efficient interpretation technique [24]. Decision trees not only can produce information about the features that a model considers important, but also, they can provide a perception of how these features interact with each other.

In computer vision, recent XAI methods have been categorized into pixel-level based and concept-level based explanations [2, 27]. Studies have shown that concept-level-based explanation methods outperform pixel-level explanations, as they overcome several its shortcomings, e.g. [2, 27, 15]). Pixel-level refers to a lower level of features such as pixels of colors, edges, or texture, whereas concept-level is a higher level of features like the nose or mouth of a human face. Additionally, several seminal works in the literature verify that concept-based explanations are more human-friendly since they can enhance the way complicated models are perceived by humans [2, 27]. At that in- stance, to predict a police vehicle, wheels or police logos are more meaningful than the vehicle's edge or surface. Consequently, the focus of XAI research has shifted from pixel-level features towards mid- or high-layer feature maps, expressed as Concepts Activation Vectors (CAVs) [27]. Despite the recent interests in the concept-level direction, most of the proposed work only target approaches to extract concepts, CAVs, as explanations. They do not discuss how end-users can eventually utilize them to understand and trust a CNN model. A small number of investigators have examined their XAI methods with reliable human evaluation studies. The lack of an agreement on interpretability definition leads to the absence of a unified mechanism of XAI algorithms evaluation [40].

One of the state-of-the-art algorithms from the concept-based explanations research direction is the Invertible Concept-based Explanations, ICE, introduced by Zhang et al. [57]. ICE is a powerful invertible tool that learns features as non-negative concepts activation vectors (NCAVs). ICE uses a non-negative matrix factorization (NMF) as a reducer for the CNN feature maps. ICE overcomes the shortcomings of the existing concept-based explanation methods and provides both local and global explanations automatically. We found that the extracted NCAVs outperformed the previous produced (CAVs) interpretability and fidelity. It is also distinguished by its capability of transforming a generated explanation back to the target model as a two-way explainer. However, ICE uses a linear classifier to approximate the original CNN model and make the predictions of their final framework.

The main contributions of this work include twofold. We introduce a novel framework, TreeICE, that modifies the original work, ICE [57] to provide a concept-based local and global explanation using a surrogate decision tree that is visualized at a semantic level. In addition, we conduct a systemic and scientific evaluation methodology to investigate our research goal and highlight the so-called trade-off. The evaluation consists of computational experiments and a human subject study. This strategy shows its reliability and can contribute to the XAI evaluation field.

The remainder of the paper is organized as follows. Section 2 presents a preparatory background followed by a review of related work. Section 3 introduces the methodology of the proposed algorithm, TreeICE. The stages of our novel approach are described thoroughly, and the generated explanations are shown. Section 4 evaluates TreeICE and discusses the findings we obtained. We compare our method against ICE, which is the original work we modify. Section 5 addresses a proper discussion to additionally analyze the research findings and connect the research parts. Section 6 summarizes this research with a related concluding discussion and recommendations for future works.

## 2 | BACKGROUND

### 2.1 | Decision Trees

A decision tree is typical a data mining model for building classification systems based on numerous variables or developing prediction algorithms for a target variable [49]. It is a non-parametric method, which means it can handle large and complex datasets without enforcing a parametric framework [49]. Study data may be split into training and validation sets if the sample size is sufficiently large. Several decision tree algorithms have been developed and arecurrently widely used, such as CART, C4.5, ID3, CHAID, and QUEST [8].

Decision trees imitate the way humans think by its naturality of stratifying samples data to sub-samples through conditional statements (e.g., if-then rules) [28]. These rules are presented in a branch-based graph that form an inverted tree with some nodes and branches. Following certain paths based on classification rules and a determined threshold value, a final prediction can be made. For instance, *"if condition* 1*, condition* 2*, and condition n occur, then outcome Y results"*.

The interpretability of decision trees enables humans to understand why a prediction or a classification is made. It also helps to show the rationale behind the choice to accept or override the model's output. All in all, the decision tree models have been characterized by their interpretability, making them one of the most sought-after algorithms when comprehending the underlying predictions is required.

### 2.2 | Explainable AI

Explainable AI or XAI term was first introduced in 2004 by [51], to illustrate the ability of their system that described how AI-controlled entities behaved in simulation games applications. While the term appears to be recently adopted, the issue of explainability dates to the mid-1970s, when investigators began studying how to explain expert systems [1]. In the following years, the research work of explainable algorithms slowed due to the revolution of AI and the focus on AI and ML systems performance [1]. However, the interest in XAI studies has recently renewed because of the widespread of complex AI systems, which impact decision making processes in various industries [1]. Miller [35] indicates that humans are naturally curious to understand and interpret AI systems' decisions. Therefore, social, ethical, and legal demands have been called for XAI. While there is an absence of a standard and general definition of XAI terminology, many prior publications debate the nuances of these different definitions. DARPA [20], for example, states that XAI aims to "produce more explainable models, while maintaining a high level of learning performance (prediction accuracy); and enable human users to understand, appropriately trust, and effectively manage the emerging generation of artificially intelligent partner".

## 2.3 | Taxonomy of XAI

Although XAI is a relatively new field requiring definite classification, XAI algorithms can be categorized broadly based on model target, range, and mechanism. Figure 1 organizes this overall classification from the literature [1, 40].

From Figure 1, it can be inferred that the intrinsic models are model-specific by definition, however, noteach model-specific algorithm is also intrinsic. Moreover, model-agnostic methods are usually post-hoc models, but not always. This introduced taxonomy is helpful to understand the state and the range of each XAI algorithm.

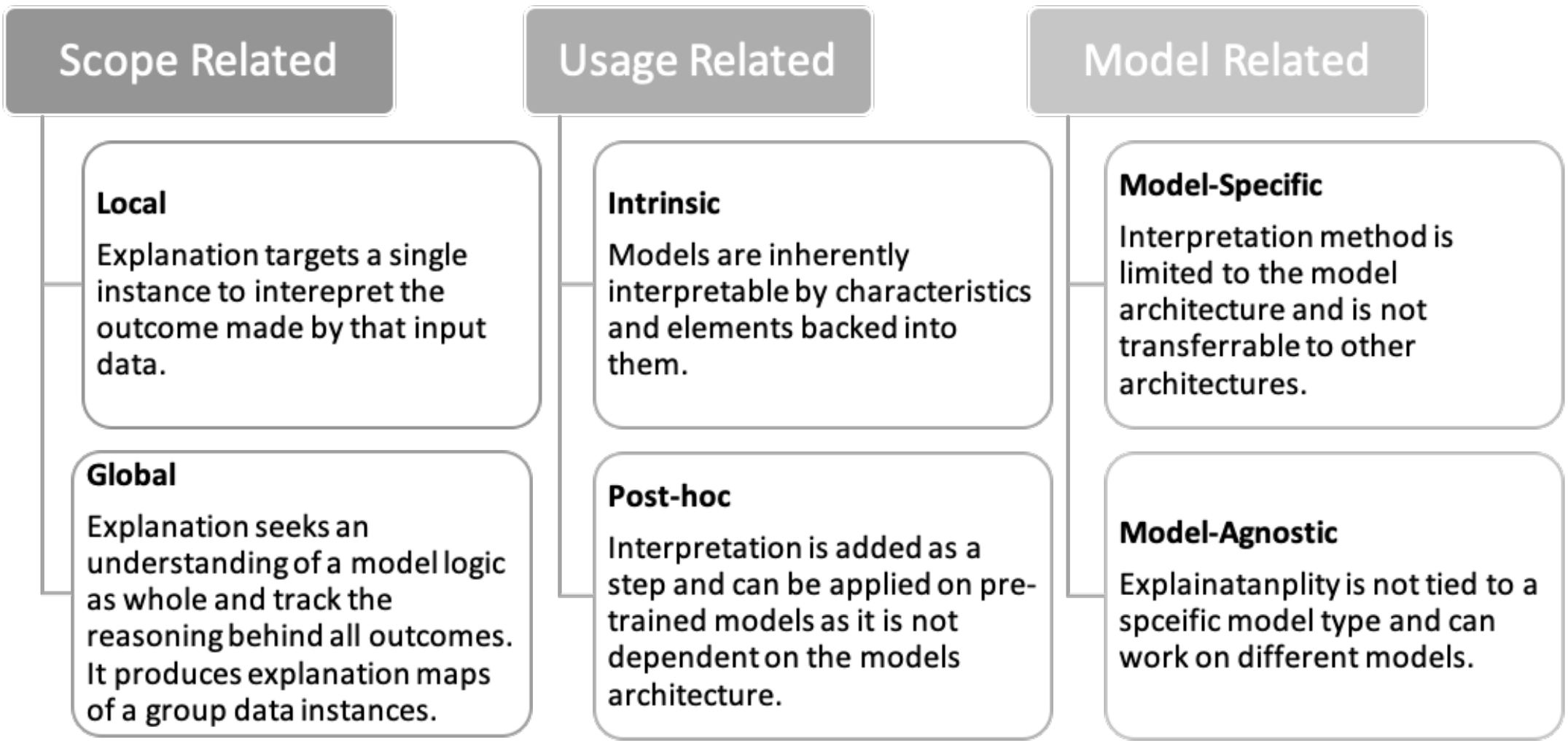

**FIGURE 1** Classification of XAI methods based on scope, usage, and model mechanism.

## 2.4 | Explainable Deep Learning in Computer Vision

Explainable deep learning (XDL) is one of the XAI directions that solely aim to explain deep learning algorithms. This review focuses on XDL approaches that target computer vision tasks, such as image classification and detection. That is, some methods that explain DNNs, particularly CNNs, for image tasks would be reviewed. The research line has started to understand DNNs by quantifying features importance at the pixel level, focusing on patterns visualization. This type of method has been known as pixel-based or input-based explanation. It generates a map to capture how important each image pixel is to its prediction label. However, the subsequent studies addressed several flaws of pixel- based explanations, e.g. [15, 17, 27, 14]. For example, Kim et al. [27] supported the fact that pixel-based explanations are unreliable when an explanation is sensitive to circumstances that do not add to the model prediction. Kim et al.
[27] also demonstrated that pixel-based explanations that do not provide input invariance, as a requisite for reliable attribution, could lead to misleading attribution. Another apparent drawback is that most of these algorithms process the explanations only locally. It has also been confirmed that these methods are subjects to human bias [27]. Most importantly, human experiments verify that these methods do not enhance human understanding and trust for the model under explanation [27, 42].

As a result, another research direction has been pursued to explore higher-level features. The goal is to learn concepts representations of DNNs layers and evaluate these concepts importance. These methods detect the important concepts instead of the important pixels. The pixel-level features are extracted from the features in the lower layers of a DNN model whereas the concept-level features are generated from higher layers.

When it comes to developing a mental model of a visual system for experts and non-expert users, concept-level features are more efficient, less confusing, and more natural than pixel-level features, according to [27]. Humans can smoothly generalize their understanding to future input data or tasks with concept-based explanations [2]. We can conclude that concept-features are more informative than pixel-features. Given these studies, we synthesize the primary taxonomy of XDL methods, on image tasks, into pixel-based and concept-based explanation methods [27, 2].

A number of seminal works that have been done in both directions are analyzed in the next subsection. In Table 1, we map the general categorization of XAI methods, from Figure 1 to the related works we reviewed of both research directions as XDL method types.

| **XDL Method** | **Sc ope** | **Usage I** | **Mode** | **Type** | **Year** |
|---|---|---|---|---|---|
| Activation Maximisation [12] | LO | PH | AG | Pixel | 2009 |
| Explainability Vector [3] | LO | PH | AG | Pixel | 2010 |
| Saliency Maps [47] | LO | PH | AG | Pixel | 2014 |
| LIME [43] | LO | PH | AG | Pixel | 2016 |
| SP-LIME [43] | GL | PH | AG | Pixel | 2016 |
| CAM [58] | GL | PH | AG | Pixel | 2016 |
| .2 Grad-CAM [46] | GL | PH | AG | Pixel | 2017 |
| 1 SHAP [32] | LO/ GL | PH | AG | Pixel | 2017 |
| TCAV [27] | GL | PH | AG | Concept | 2018 |
| Anchors [44] | LO | PH | AG | Pixel | 2018 |
| Zhang et al. [56]'s algorithm | LO | PH | SP | Concept | 2018 |
| ACE [15] | GL | PH | AG | Concept | 2019 |
| CoCoX [2] | GL | PH | AG | Concept | 2020 |
| CNN2DT [24] | LO/ GL | PH | AG | Concept | 2020 |
| ICE [57] | GL | PH | AG | Concept | 2021 |
| ACDTE [11] | GL | PH | AG | Concept | 2021 |

**TA B L E 1** Summary of the reviewed explainable deep learning algorithms (XDL), image-related tasks, along withthe corresponding categorization. LO: Local, GL: Global, PH: Post-Hoc, AG: Model-agnostic, SP: Model-Specific, Pixel: Pixel-based explanation, Concept: Concept-based explanation

## 2.5 | Related Work

In this section, we present an in-depth review of the related work on explaining image classification. We limit our review to the research that are applicable to explain CNN models. Some of these works will be introduced with moredetails considering their importance in the research community.

## 2.6 | Pixel-based Explanations

A pixel-level explanation is an explanation extracted from the lower-level features. We select a few of the foundationalworks in this research direction and analyze them as follows. Erhan et al. [12] presented Activation Maximization as one of gradient-based methods on deep architectural visualization. The work visualizes the important features in any layer of a deep architecture by optimizing the input to maximize the activation of selected unit in a target layer. One of issues in this method that it provides only local explanation at pixel level.

Baehrens et al. [3] introduced the idea of explainability vector technique to explain arbitrary nonlinear classification outputs. An explainable vector can be defined as the derivative of the conditional probability data of a class given by a Bayes classifier. The work simulates a complex classifier in a local area by training a Parzen window classifier that is similar to a Bayes estimator, so that the explanation vectors can be estimated. This method can be known as a gradient-based algorithm that describes how a change on an input image would be responsible for its prediction class.

Simonyan et al. [47] proposed one of the notable works on saliency map technique that shows how calculationsof input image gradients can reflect the influence on the image's output class. This work produced two visualization methods for CNNs, a class model and an image-specific class. The first one generates an artificial image as a classof interest representatives, and the latter highlights the regions of the input image that are discriminative to the assigned class. Although saliency maps and their extended variations and improvements have provided relatively good visualizations, it suffers from several drawbacks. For instance, they are not class-discriminative since the visualizations of different classes are approximately alike. Furthermore, Ghorbani et al. [14] showcased that saliency maps could be vulnerable to adversarial attacks.

To mitigate the class-discriminative limitation of saliency map methods, Zhou et al. [58] introduced Class Activation Mapping (CAM) algorithm. CAM intended to limit the class-specific image areas with a single forward-pass to visualize the predicted class on any image. The method adapts the global average pooling function (GAP)to improve CNNs visualization. Although this approach is highly class-discriminative, it trades off the high performance of a CNN model to obtain more transparency. As a result, the Grad-CAM, Gradient class activation mapping,technique generalized the CAM method to overcome its shortcomings [46].

Grad-CAM proposed a class-discriminative localization system that does not alter CNN models architectures. It not only evaded the explainability versus accuracy dilemma but also extended the CAM to cover more CNN modelfamilies and provide better visualization. However, as Grad-CAM is a gradient-based method, it suffers from some gradient issues. For instance, Kim et al. [27] showed that that most gradient-based algorithms improperly assign constant vector transformations and that input invariances should be required for accurate attributions.

Local Interpretable Model-Agnostic Explanations (LIME) [43] is one of the most popular local approximation methods that intends to produce an intelligible and understandable representation. LIME attempts to determine the importance of continuous superpixels (a patch of pixels) in image data with respect to the output class.

LIME finds a binary vector to identify the existence or absence of a continuous path, or superpixels, that gives the best representation of a class label. However, LIME has a flow with the precision and coverage as they may not be assured. Considering LIME explanations are formed locally in linear way for new instances, these instances could be outside the area of the linear combination of input features. Thus, the linear and local explanation may not remain possible. Several seminalworks have been introduced following LIME in order to extend it and improve it. For example, SP-LIME is presented by Ribeiro et al. [43] to extend LIME and provide global explanations instead of having only local explanations.

Shapley Additive Explanations (SHAP) is a pixel-based method developed by [32]. It is a game theoretically optimum approach based on Shapley values for model explainability. As with LIME, the explanations are feature contributions to the predicted output. Shapley values may be estimated to identify how to divide the payoff equitably by treating the data features as participants in a coalition game. Like LIME, a data feature in the SHAP technique may represent individual categories in tabular data or superpixels groups in image data. SHAP has been an active topic in the research community and a number of extended works have been investigated to improve the method.

## 2.7 | Concept-based Explanations

Concept-based explanations have been explored because of the limitations detected in the pixel-based explanation approaches. Concept Activation Vectors (CAVs) notion has been first introduced by Kim et al. [27] using Testing with Concept Activation Vector method (TCAV). TCAV produced a framework to extract concept-level explanations using high- level concepts defined by human. That is TCAV is a new linear conceptual interpretability approach with quantitative Testing to provide a global explanation and shift from pixel level methods to concepts level methods. TCAV employed directional derivatives to measure a model's prediction based on the extracted CAVs. One of the most significant shortcomings of this algorithm is that it experimented with a pre-defined concepts queried by humans. Although it is helpful when users have a collection of pre-provided concepts and sources to present examples, it is a concerning obstacle as the range of these concepts may be ambiguous or unbounded. CAV also depends on human bias during the explanation procedure since users may choose concepts incorrectly.

Ghorbani et al. [15] presented a novel unsupervised learning approach for TCAV, named Automated Concept Explanation (ACE). It aims to automatically generate only global explanations without human supervision. *k*-means Clustering [13] was adopted in ACE to extract a set of visual concepts. These concepts are represented as vectorsderived from cluster centroids. The generated concepts by ACE were demonstrated as human-readable concepts thatare also deemed critical to the corresponding neural network. However, concepts extracted by ACE are not always faithful. They lack fidelity compared to other reduction methods, such as Principal Component Analysis (PCA), as shown in [57]. Another flaw of ACE algorithm is that a significant amount of information could be lost when invertingthe inputs back to the original dimension due to the use of a hot vector to store information.

Another seminal work explored by [2] is CoCoX, which stands for Conceptual and Counterfactual Explanations. CoCoX extracts automatically explainable concepts from a provided training dataset and generates fault-line explanations. Using Grad-CAM algorithm [58], it analyzed feature maps from the last layer of a CNN model as examplesof concepts and took their localization maps. Then, applying directional derivatives from the TCAV method [27], the importance of the concepts was computed. The derivation of fault-lines was then formulated as an optimization problem to choose a minimal collection of these concepts that alter the model's prediction. The evaluation study of CoCoX also showed how it outperforms several competitive baselines in the field. However, CoCoX suffers from an issue that most of the proposed concepts-based explanations have. The work focused on extracting understand- able concepts, but it did not discuss how an end-user would finally use them as an explanation tool.

Invertible Concept-based Explanations (ICE) was developed by Zhang et al. [57] to modify the work ACE [15] and overcome some of its considerable shortcomings. ICE generates both local and global concepts-based explanations, using a non-negative factorization matrix (NMF) reducer instead of *k*-means as a clustering method. Zhang et al. [57] showed that the dimensionality reduction methods perform better than segmentation and clustering methods. ICE showed that NMF evaluated information loss in the explanations using inverse functions, which enables analyzing the contribution distributions from the derived NCAVs to produce detailed local explanations. This inverse function enables the ICE method to invert and have two-way explanation system. Lastly, ICE employed a number of XAI evaluation metrics to measure fidelity and interoperability in comparison with the ACE model and another popular matrix factorization method, Principal component analysis (PCA). Using human behavioural experiments, the authors showed that ICE is superior to ACE as it produces concept-based explanations that are more interpretable and faithful.

## 2.8 | Decision Tree and Neural Networks

The distillation of knowledge from neural networks into tree structures has been an active research trend. For instance, TREPAN was introduced by Craven and Shavlik [9] in 1995 as one of the earliest works that integrates a decision tree and a neural network. TREPAN aimed to build a decision tree that approximates the concept represented by aparticular neural network.

One of the recent works in this direction is the algorithm presented by Zhang et al. [56] to learn a decision tree that explains the reasons for each prediction made by a CNN model at a semantic level. Their decision tree decomposed the feature representations of high-level layers into concepts of object parts/filters. Therefore, it helped to understand quantitatively how much each filter contributes to the prediction score. The algorithm derived all possible CNN decision modes, a decision mode refers to a case of how the CNN makes predictions using filters. However, this method has an apparent shortcoming as it implies retraining the whole network to have each filter detects a particular concept before learning a decision tree. It also lacks a proper human or user study to validate the interpretability of the work efficiently.

Jia et al. [24] proposed CNN2DT as a method for training surrogate decision trees from CNNs and developing an exploration-oriented visual analysis system. The results demonstrated that CNN2DT interprets the decision-making process of CNNs with global and local explanations. The network dissection method [4] was employed to map each feature to human-readable semantic labels, so the decision tree would be tangible to humans. On this basis, this work presented CNN2DT (Convolutional Neural Network to Decision Tree) as a visual analysis system for interpreting the decision process of a CNN model utilizing procured surrogate decision trees while maintaining interpretability in a way comparable to human decision-making. One of the limitations in this work is using a labelled semantic datasetinstead of generating these semantic labels automatically. Another issue is that the structures of the learned decision trees were considered unstable when using various amounts of train instances.

Automated Concept-based Decision Tree Explanations (ACDTE) is another research by [11] to extract concept features and employ a decision tree model that approximates an original CNN model and generates final explanations. ACDTE identified a group of images comparable to the image to be explained from either the primary task dataset or a secondary dataset. A shallow decision tree was trained using the feature vectors and the model's predictions. El Shawi et al. [11] argue that ACDTE provided not only a counterfactual explanation but also a natural explanationfor the relevant concepts to the prediction. Even though it considered extracting the concepts automatically, it used clustering methods which NMF reducer showed its superiority in terms of interpretability and fidelity [57]. Another important point is that this work trained a linear model to act as a binary concept detector classifier before using their final decision tree model for explanations. This is because the clustering algorithm finalizes its predictions of theextracted concepts in the form of binary labels, 0 or 1. Thus, there is a need to predict the existence of the derived concepts in each input image. Having this linear model adds some performance issues that also require tuning andevaluation.

## 3 | METHODOLOGY

We aim to build a usable concept-based explanation framework that can provide an interpretable and faithful system. Considering the seminal works discussed in Section 2.8, not only does the decision tree model reveal information about the important derived features, but it also provides a sense of how they relate to each other. We, as a result, believe that training a decision tree with the concepts generated by ICE would yield a robust explainable and usablemodel for a target CNN model.

### 3.1 | TreeICE Framework

We develop a concept-based explanation algorithm that describes the underlying features of the complex CNN learner. At a glance overview, we extract the concept-level features using the ICE algorithm. These concepts are then fed to a decision tree model and classified into their belonging classes. The final output is a generated explanation in a tree form that shows how a classification decision has been made and how the derived concept features relate.

The main stages of our approach are that the pre-trained CNN model which works as a concept extractor and the decision tree approximation which serve as a concept classifier and an explanation. Non-negative matrix factorization (NMF) is applied as a reducer on the CNN model's feature maps to produce Non-negative Concepts Activation Vectors (NCAVs). Thus, NCAVs are extracted automatically without the need to pre-labelling them manually. A target dataset is used to train the NMF. Each generated NCAVs would be ultimately represented as a score that indicates its importance to the associated concept. We use the extracted NCAVs as data points to train the decision tree model. To form the training set of the decision tree model, we propose using either the CNN model predictions or the ground-truth labels with the corresponding extracted NCAVs. That is, our framework can be used in two modes. If the CNN model's predictions labels are selected, our method with a surrogate decision tree can be considered as anexplainable model that imitates and interprets the original CNN model. If the ground-truth labels are trained, the CNNmodel can be thrown away, and our proposed method can directly be used as a standalone model. The selected label set is stored along with the corresponding NCAV scores to train TreeICE model. A new input can also be predictedusing the same process and after learning concepts from its feature maps. Figure 2 provides a diagram that describes the proposed framework.

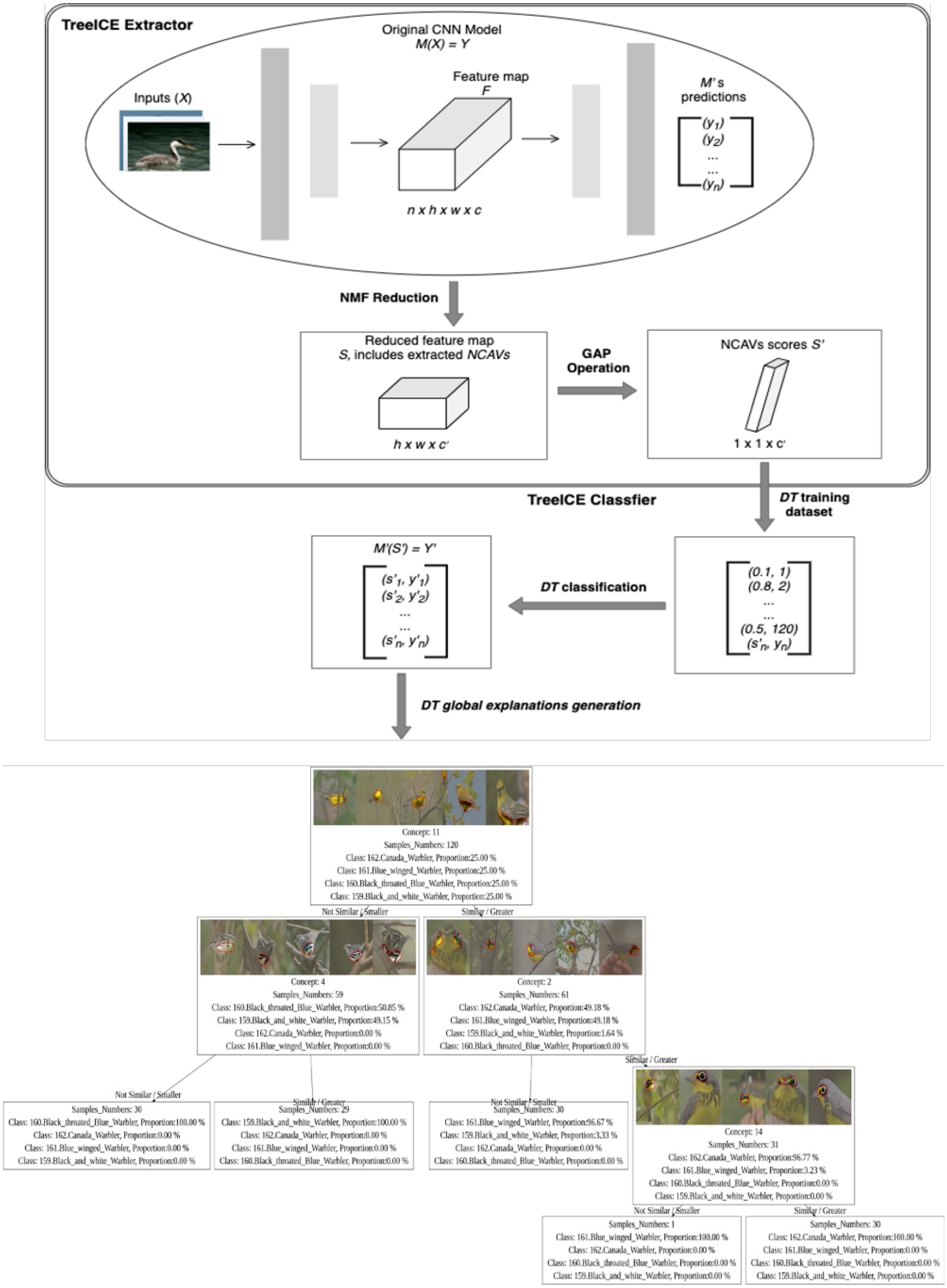

**F I G U R E 2** Diagram of the TreeICE framework to show the main stages of the proposed tree approximation. TreeICE extractor is the first stage which extracts the concepts, NCAVs, and apply GAP process to create final scores of each NCAV. The second stage is TreeICE classifier where NCAV and their corresponding labels form a training dataset. The labels (y) can be either the ground-truth labels or the prediction labels by CNN model (M's predictions). This dataset is used to learn a *DT* classifier and generate a tree-like explanation.

## 3.2 | Concept Extractor

This first stage of our algorithm is to extract the concept-based features. We start with a description of the methodology's notations and setup. Given a pre-trained CNN model $M$ and $X$ input images, a prediction of an instance can be processed as $M(x_i) = y_i$, $1 \leq i \leq n$, where $n$ is the number of input instances, $x \in X$, and $y \in Y$ denotes the scalar classification score after removing any softmax layers. A target layer $l$ has to be selected to get the feature maps $F$. According to [55], we can learn high-level features (concepts) from higher layers while lower layers intend to be for lower-level features such as edges or textures of an image. Therefore, in such work, the selection of a suitable target layer is extremely important. As ICE algorithm suggests, the last layer is the best layer for concept-based explanations [57]. Let $F$ has the shape $n \times h \times w \times c$, $h$ (height), $w$ (width) both refer to the size of $F$, and $c$ is the channel dimensions or the number of the feature maps. If the CNN model uses the *Relu* activation function, $F$ is guaranteed to be a non-negative matrix. This will be needed to be able to use NMF reducer properly. We consider $v$ is a vector in $F$ at a position ($a$, $b$), with ($0 \leq a < h$), and ($0 \leq b < w$). $v_{a,b}$ can be a vector description of original images with different correlated receptive fields. To sum up, the input of this first phase is the CNN model and training data points, and the output is the extracted concepts denoted $NCAVs = \{nc_1, nc_2, ..., nc_n\}$. Additionally, classes predictions $M(x_i) = y_i$ are stored to be used as the objective output for training the framework's classifier. The following subsections details the steps of this stage.

### 3.2.1 | Non-Negative Concept Activation Vectors (NCAVs) Extraction

Non-negative matrix factorization (NMF) is distinguished by its ability to automatically extract sparse and importantfeatures from non-negative data vectors [16]. The objective of NMF is to find two matrices, $W$ and $H$, of the original matrix $V$ that contain only non-negative elements. NMF then implies that the data inputs have a set of hidden featuresexpressed by each column of $W$ while each column in $H$ denotes the coordinates of a data point in $W$. $W$ column can be described as basis of images and $H$ column plays a role of showing how to sum up the image basis to extract anapproximation of a particular image. Each input of $V$ is approximated using an additive combination of non-negative vectors.

The ICE method [57] leverages the use of NMF to derive concept-level features and design a robust concept- based explanation framework. They employed NMF to produce Non-negative concept activation vectors (NCAVs) from a CNN model feature map. To elaborate using our settings, having $F$ flattened to $A \in R^{(n \times h \times w) \times c}$, the NMF reduces $c$ to $\hat{c}$. We can say now $A$ is reduced to $S$ and $D$ when $A = SD + U$. Considering the example of NMF reducer we illustrated, $V$ can represent $A$, $W$ refers to $S$ and $H$ is $D$. $S$ denotes to NCAVs scores as $S \in R^{(n \times h \times w) \times \hat{c}}$ while $D$ is the feature direction and $D \in R^{c, \hat{c}}$. To effectively apply NMF, the residual error $U$ needs to be minimized, as equation1 formulates [57] :

$$U = min||A - (S \times D)||_M, S, D \geq 0 \quad (1)$$

The vector $v$ of images from the same class that share similar concepts can be presented as a vector description of a target label part. For example, the red-eye of a bird from class x can be a concept-level feature, so all images with red eyes represent the vector description of class x. Using NMF to factorize these vectors can disassemble significant and most related directions of the target class. To be more precise, $D$ is a fixed parameter that we get after training NMF reducer on a target dataset. $D$ is themeaningful NCAVs of feature maps in the assigned dimension space. It is used to extract concepts from new instances and generate a local explanation. As $D$ is a vector basis and $S$ is the length of $v$ estimations on $D$, $S$ is the score that represents the similarity degree of $v$ to NCAVs in $D$.

### 3.2.2 | Applying a Global Average Pooling Layer

Pooling is a primary operation in recent CNN architectures to reduce dimensionality and complexity [5]. One of the commonly known pooling methods is Global Average Pooling (GAP). In this work, we apply GAP as a mean filter to down sample each feature map and convert its values to a single number. Thus, we transform $S$, the extracted NCAVs feature maps, from the shape ($h \times w \times \hat{c}$) to ($1 \times 1 \times \hat{c}$). We note that the depth $d$ from the GAP example refers to the number of channels $c$ in our settings. It is important to point that GAP is a fundamental step in our algorithm as it helps to draw the final NCAVs as scores $\hat{S}$ in one-dimensional size so that they can be prepared and fed to a classifier properly without any further transformations.

### 3.2.3 | NCAVs Visualization

Visualizing concept vectors is necessary in order to understand the underlying features of a complex system. In this work, we follow the ICE algorithm's visualization technique to present the extracted concepts [57]. ICE employed the prototyping method produced by TCAV [27]. The intuition behind the TCAV prototypes is to collect images instances that have the desired concepts and emphasize these concepts. The highest concept scores of $\hat{S}$ that were generated after GAP process are used to determine a prototype's instances. Hence, five images with the highest scores of a particular concept are chosen as samples to represent that concept. Reduced feature maps $S$ of a given NCAV can be a heatmap for a corresponding concept. In the original ICE work [57], a heatmap and an image are integrated, and a threshold of 0.5 is set to determine areas with high concept value in an image. After a minmax normalization, areas of an image with values higher than 0.5 are highlighted. Figure 3 illustrates an example of to a visualized concept (Eye). To have our ultimate tree-like explanation in its semantic visualization, we use the *GraphViz* (https://graphviz.org/).

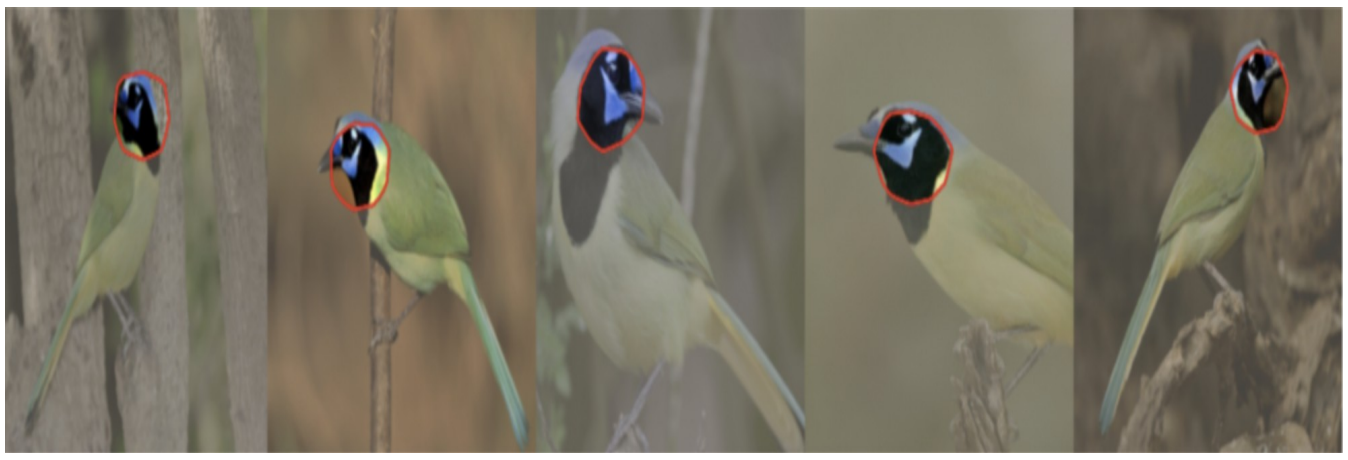

**F I G U R E 3** An example of a concept extracted (Blue-black eye) that is represented using 5 images from the dataset as a prototype of the concept.

## 3.3 | Concept Classifier

The second primary stage in the TreeICE approach is the classification of the extracted concepts to their corresponding categories. The input of this phase is a dataset that yields from the first stage, the learned NCAVs scores $\hat{S}$ along with $M$'s predictions ($Y$). For each $\hat{s} \in \hat{S}$, we can formulate the training dataset as $G_{ncav} = \{ (\hat{s}_1, y_1), (\hat{s}_2, y_2), .., (\hat{s}_n, y_n) \}$. This stage output is the prediction labels of our TreeICE explanation as that $\hat{M}(\hat{S}) = \hat{Y}$.

### 3.3.1 | Learning Concept Decision Tree

The goal is to build a decision tree (*DT*) to have a tree approximation that mimics the behavior of the original model, CNN. *DT* has been elected as our main explanation due to its interpretable nature that enables the derivation of concept rules through a root-leaf rout. It contributes to the generation of knowledge on the underlying features of complex models. In this case, it can provide more details about the extracted concept vectors to comprehend a CNN model. Thus, $G_{ncav}$ is used as input to train *DT* model which is responsible for classifying the NCAVs instances and produce the final prediction labels of our TreeICE explanation. We use an optimized version of the CART algorithm [6] was used based on the *Scikit-learn* implementation [41] to train the *DT*.

## 3.4 | Generated Explanations

The TeeICE framework can produce both global and local explanations. We show these two explanations in the examples below to illustrate the final produced explanations by our model. The numbers of classes and channels (concepts) are first selected. To simplify the idea, we train TreeICE with only four classes and 10 concepts. A decision tree consists of nodes that undertake tests on the selected features. To check if a variable has a value lower than, equal to, or greater than a pre-defined threshold. Each branch indicates a possible decision, and leaf nodes refer to a class label [19]. Thus, a path from the root to a leaf represents the classification rules. In our model, each node in the tree contains one concept (feature), depicted with five image samples, as well as some information to help understand the decision path. However, the most important information for non-expert users is the visualized concept (the feature). Therefore, we intend to show most of the related information for expert users, such as the number of samples and the threshold value. The number of samples refers to the number of training data points used at the node, while thethreshold value indicates the value set by the classifier to split the node. Each branch decides whether a starting node has a concept that is Similar or Not Similar to the concepts of a trained class. The visualized concept and the shown information in each node are adequately illustrated as the semantic classification rules of the decision treethat explain a decision made.

### 3.4.1 | Local Explanation

The local explanation justifies the prediction of a given instance or test image. Each extracted concept of the test image is related to the most similar NCAV, which had already been generated when the NMF reducer was trained. Figure 4 gives an example of TreeICE local explanation for the prediction of a test image. In this example, TreeICE istrained with 4 classes, and the number of channels (number of extracted concepts) was reduced to 15. We highlight, with a blue color, the path of the decision made for the test image. The classifier makes the decision that the test image belongs to Class 162 (Canada Warbler). This colored decision path is chosen as the classifier compares towhich degree the input image is similar to each extracted concept from the four trained classes. For instance, in the first node, the highlighted concept 'Concept 9' is a yellow body, so the tree makes the decision Similar as the test image has a yellow body too.

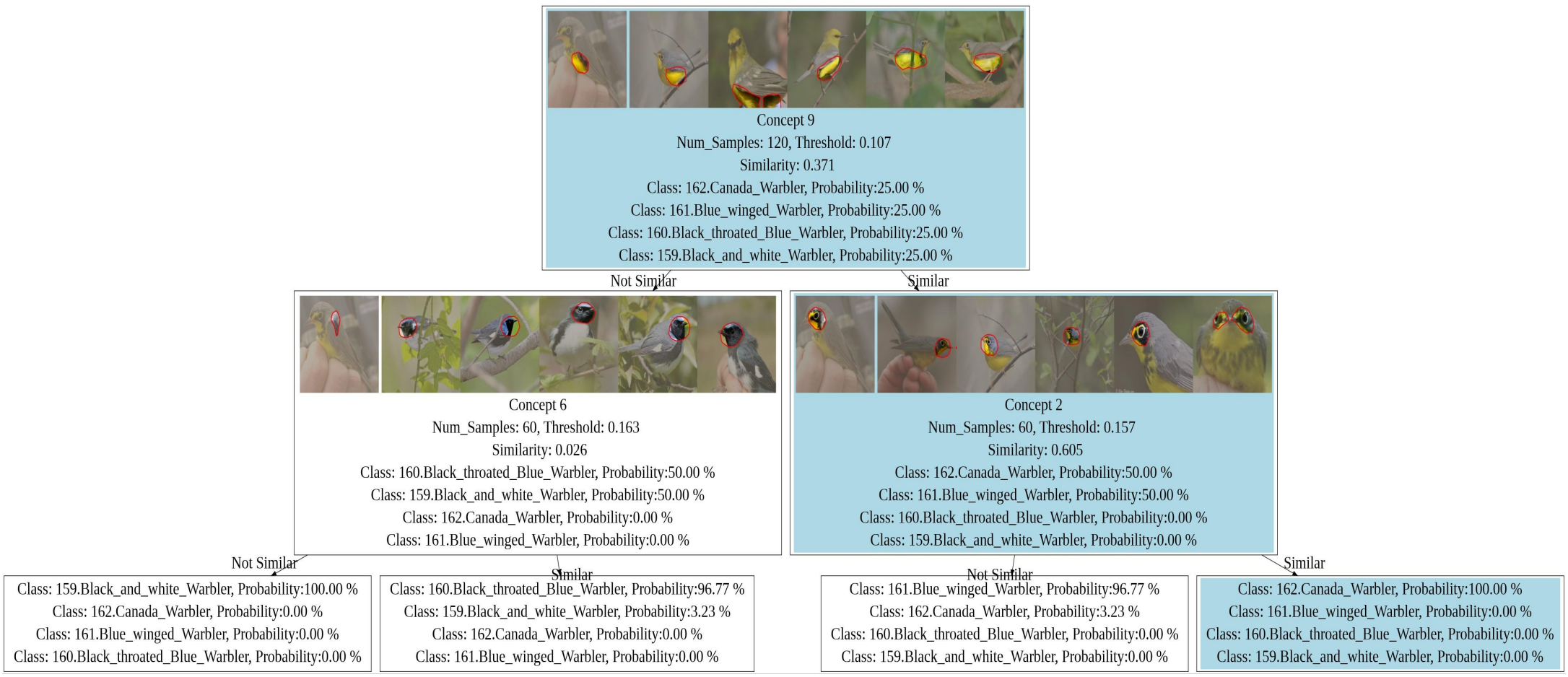

**F I G U R E 4** Local explanation generated by TreeICE. The colored path shows the decision path of the explanation model.

## 3.4.2 | Global Explanation

The global explanation provides a holistic view to describe all the classes of the target dataset. Figure 5 presents an example of the global explanation generated by TreeICE. As it can be seen, each node has an extracted conceptand selected some information. The leaves nodes of the global explanation indicate a possible decision outcome of each trained class with a corresponding path. For example, a decision path shows that 'Concept 2' and 'Concept 8' together are the conditions of Class 40 (Olive-sided Flycatcher) with a probability of 100%. Thus, the path explains this decision as that Class 40 is recognized with these two concepts among the trained classes.

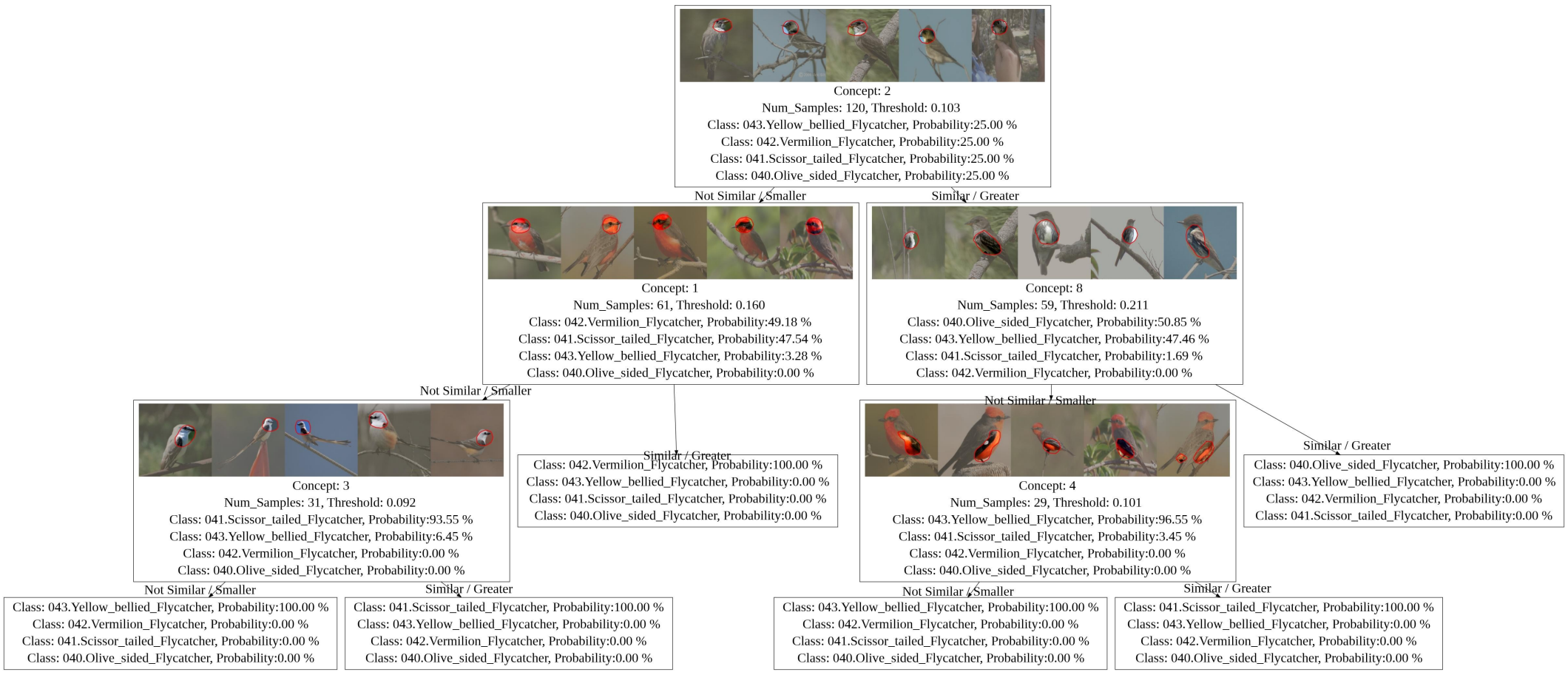

**F I G U R E 5** Global explanation generated by TreeICE, our proposed method.

# 4 | EVALUATION

To efficiently assess the proposed work, TreeICE, we conduct both computational and human subject experiments for empirical evaluation. Since we investigate an XAI framework that aims to produce an explainable model, we adoptthe necessary desiderata: fidelity, performance and interpretability, synthesized by Guidotti et al. [19]. To validate the fidelity and the performance of an explanation, we use computational evaluation [31]. Inspiring by [23, 10], we employ human subject experiments to measure the interpretability.

We aim to evaluate the tree approximation we adopted as an explainable model. Therefore, we compare our alternative framework, TreeICE, against the algorithm ICE [57], where they used a linear approximation. We re- implemented their work and used it as our main baseline for our experiments. In this section, we refer to the baseline ICE as Linear model (*LR*) and our model TreeICE as Decision Tree model (*DT*). Although our framework can be applied to several image processing tasks, the image classification task was chosen for its simplicity and importance.

## 4.1 | Evaluation Setup

To organize our evaluation for both computational and human subject experiments, we divide the design into main components as: 'Independent variables', 'Dependent Variables', 'Constants', and 'Metrics'. We first define only the 'Independent variables' and 'Constants' as they are shared for the two evaluation experiments. The remaining components are described in each experiment separately.

Both frameworks, TreeICE and ICE, serve as explanation algorithms to interpret a black-box CNN model. Thatis, the explanation algorithm is the independent variable which has two values: Linear model (*LR*) as baseline, and Decision Tree model (*DT* ), as our proposed model. Thus, we have two independent experimental conditions.

The evaluation's constants play a role as infrastructures of the work and are used for evaluation experiments. Those constants are the original (black-box) model, a target layer, and a dataset. The pre-trained model *M* we targeted as an original model to be explained is the well-known CNN model for image classification tasks, introduced by [22]. Inspiring by the work we modify, ICE [57], we used the ResNet50 architecture as it shows a good result with the NMF reducer. The top1 error of this pre-trained model is 15.81%. We reproduced CNN model after tuning based on ImageNet pre-trained weights [57]. *PyTorch* [39] and *Scikit-learn* [41] were used to prepare our infrastructure. We follow the ICE algorithm [57] and choose the last layer (*layer 4*) of the CNN model as a target layer to extract the feature maps and train the NMF reducer. Furthermore, the Caltech-UCSD Birds-200-2011 (CUB-200-2011) dataset
[52] was used for both evaluations, computational and human subject experiments. It is an image dataset with 200 bird species. It consists of 11,788 bird images, 5,994 for training and 5,794 for testing. This dataset was selected as it is specialized in fine-grained visual categorization tasks. This dataset can be advantageous in our experiments since not many people are experienced with bird identification, but concept-based explanations may be understood by non-expert users. Therefore, testing our proposed method, on such complicated data can mimic real-world situations. Popular datasets with familiar images would fall short of this goal. Both models, *LR* and *DT*, were trained using the training set and evaluated on the testing set of this data.

## 4.2 | Computational Experiment Design

The dependent variables we adopted to evaluate the independent variable (the explanation method) are fidelity and classification performance. We measure fidelity and classification performance in the computational experiment.

Fidelity aims to estimate the faithfulness degree and the agreement of the produced explanation with the black-box learner [19]. It examines the correctness of the explanation model's predictions using the original model's predictions as the target labels. According to Papenmeier et al. [38], the accuracy level of the overall predictive system significantly influences user trust. Classification performance evaluates the models' accuracy, and it is defined as how correctly a model can predict unseen data points [19]. In this evaluation, we contrast the fidelity of the explanations produced by a linear model $LR$, as the main baseline, with our explanation from the decision tree model $DT$. To provide a full view of the classification performance, we evaluated comparably these the models $LR$ and $DT$, as well as the performance of the original CNN model.

To measure the dependent variables, we assign the metrics *accuracy score* and *F1-score* [19]. However, the predic- tion labels for each of them is different. The fidelity was tested with respect to the original model's predictions while the ground-truth labels were used for the classification performance. It should be noted that classification models target only the prediction labels, so the errors that do not affect these labels are disregarded.

The vast majority of Machine Learning research use the *accuracy score* to measure the performance of classifiers [48]. The *accuracy score* is defined as the ratio of the correct labelled instances, both true positives and true negatives, to the total number of input data [50]. Although *accuracy score* is one of the most applied measures to evaluate the classification performance, it has some flaws that negatively impact the outcomes in some situations. For instance, it does not differentiate between the correct predictions of different classes [21]. Therefore, we employ *F1-score* that is also widely used in the literature to evaluate the performance. *F1-score*, also known as *F-measure*, refers to the harmonic mean of precision and recall. *F1-score* can be better in multi-classification problems or when the aim is to balance between precision and recall. Given the original pre-trained image classification model $M$, $\hat{M}$ represents the approximate explanation mode and $X$ is a set of images to be classified. The fidelity of the explanation models can be defined in Equation 2:

$$Fid = \frac{\#(x \in X | M(x) = \hat{M}(x))}{\#X}$$

The parameter $c$ refers to the number of channels reduced by the NMF reducer or simply the number of extracted concepts. $k$ is the number of classes we trained the explanation frameworks with. The $DT$ model was trained with respect to the original CNN model predictions labels to measure the fidelity. However, to test the classification performance, we had to separately train and test the $DT$ model with ground-truth labels. This step was not involved when we reproduced the baseline $LR$ and used it for the evaluation. The weights of the pre-trained CNN model weredirectly used to generate linear weights, and no training process was required.

The depth of decision tree classifiers, max_depth, is a major factor that is often adjusted to inhibit overfit-ting and optimize the performance [30]. We first prioritized assessing the tree depth to obtain a clear insight of some good depth values we could use to train the $DT$ model. We assumed for the other parameters' values as that $c$ can be 15, and for $k$ we went with 10. For the maximum depth values, we experimented values from 2 to 24, in step of 2. After having a good idea about the tree size and depth, we ran both the linear and the tree explanation models with these two parameters, $c$ and $k$, to analyze how they could affect the fidelity and performance of the models. Therefore, $c$ was evaluated from 5 to 30, in step of 5, and $k$ was assessed from 4 to 14, in step of 2.The $DT$ model was also tuned using some other hyper-parameters such as, the minimum number of samples at a leaf node (min_samples_leaf), the minimum number of samples to split internal nodes (min_samples_split), and the randomness of the estimator (random_state) as trails to improve the performance of the $DT$ [41]. A grid search approach was run to select the best values of each hyper-parameter. The $DT$ was slightly improved with these added hyper-parameters.

To efficiently understand the affect of parameters, we ran 10 different groups of classes for each trial with all the assigned settings and parameters. Then, we took the average values for each aspect, fidelity and classificationperformance. For the tree depth, we also run 10 different groups of classes and get the accuracy of each group.The average accuracy of the 10 groups was computed to provide the final averaged performance of the maximum depth parameter. The evaluation experiments required around 60 hours (p100 GPU), this experiment was undertaken using the (LIEF HPC-GPGPU Facility) hosted at the University of Melbourne. This Facility was established with theassistance of LIEF Grant LE170100200.

### 4.2.1 | Results and Analysis

To generally compare the result of the fidelity and accuracy for both frameworks, *DT* and *LR,* we evaluate them with respect to the two main parameters, *c* and *k*. Analyzing the relationship of these parameters would aid in comprehending the effectiveness of the corresponding framework.

### 4.2.2 | Decision Tree Depth Analysis

The outcomes of analyzing the maximum depth of the decision tree classifier, given the settings described above aredepicted in Figure 6.

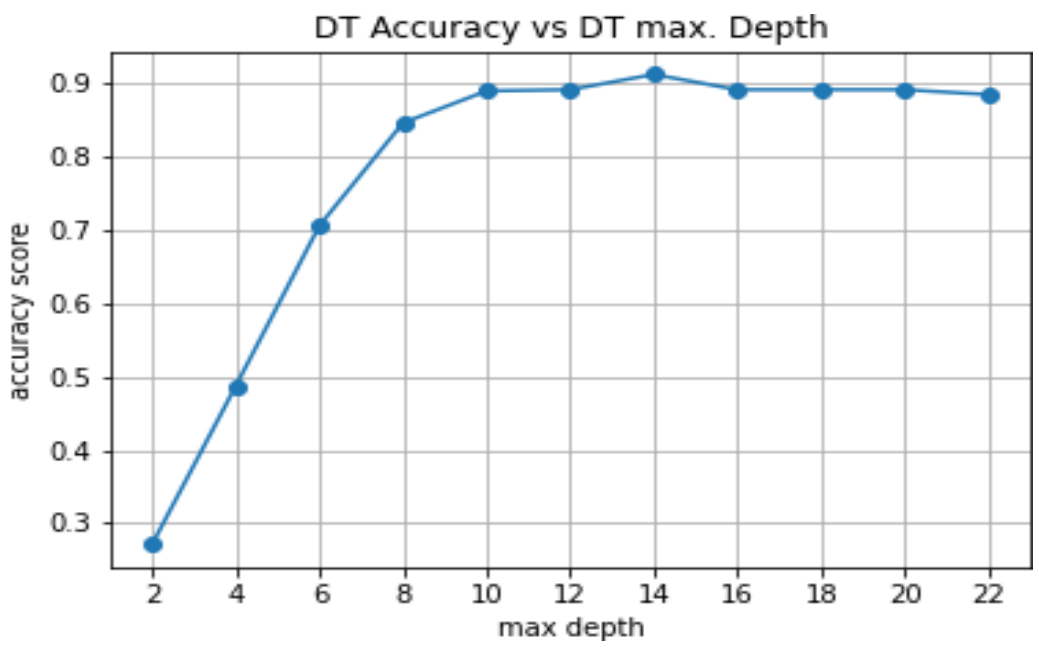

**FIGURE 6** TreeICE accuracy with different values of maximum tree depth using CUB dataset.

As it can be seen in Figure 6, the performance (accuracy) of the decision tree considerably improves as the max- imum depth value is getting greater. However, the performance roughly levels off as the depth grows beyond 10. These findings support the nature of the decision tree models since increasing in the depths refers to involving more features and rules, so the model's performance increases. These results helped us to make the decision of trainingour *DT* with $max_depth$ = 10 for the following experiments.

### 4.2.3 | Fidelity and Classification Performance Results

As we have two parameters *c* and *k*, we report the output of each model evaluation separately in 3D plots. Weobserved that both fidelity and classification performance results exhibit similar trends.

Figures 7 and 8 view the fidelity results using the *accuracy score* metric of *LR* and *DT* explanation models, respectively. For the fidelity with *F1 measure*, Figure 9 shows *LR* explanation model results while Figure 10 represents *DT* explanation model.

For the classification performance, the CNN model accuracy results along with our two competitive explanationmodels are assessed. Figures 11, 12, and 13 illustrate the performance outcomes with the *accuracy score* metric whereas Figures 14, 15 and 16 refer to the classification performance using *F1 score*.

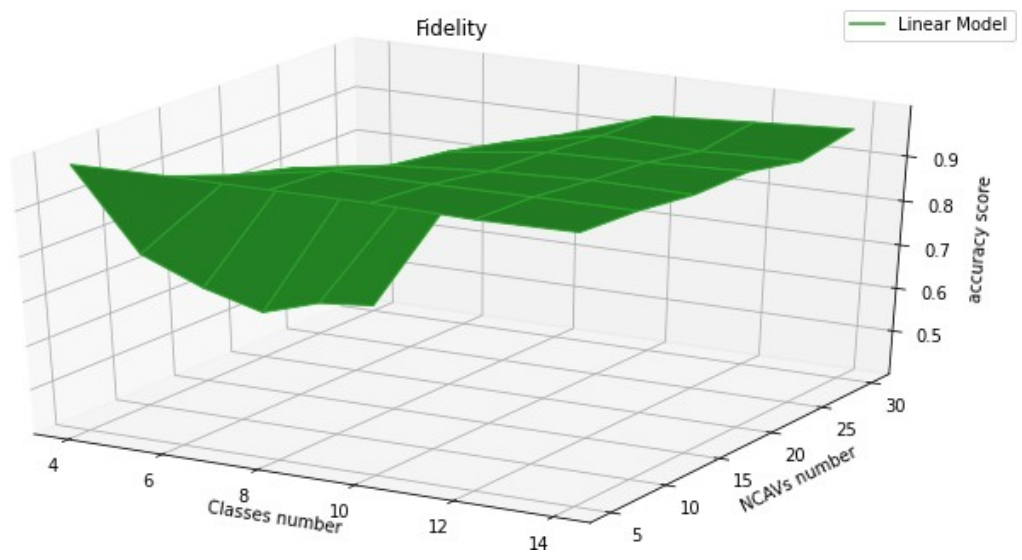

**F I G U R E 7** *LR* model fidelity scores, using *accuracy score*

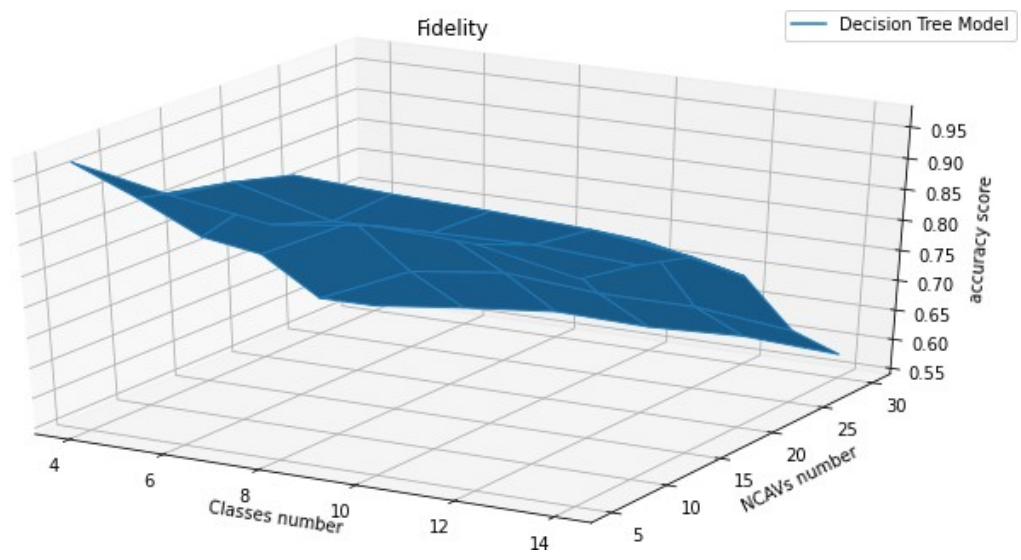

**F I G U R E 8** The fidelity of *DT* model, using *accuracy score*

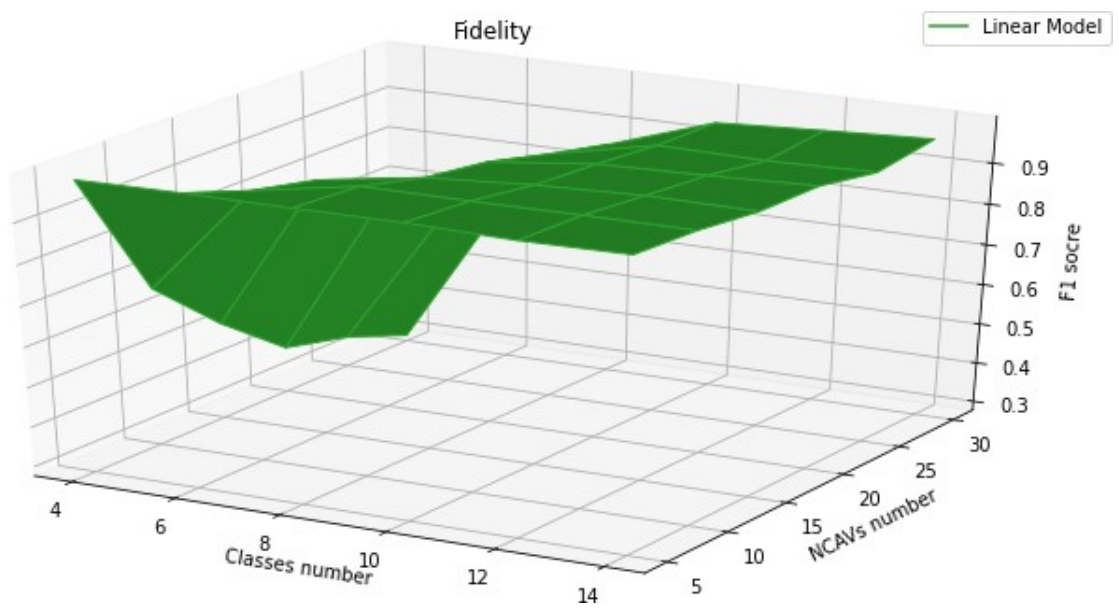

**F I G U R E 9** *LR* model fidelity scores, using *F1 score*

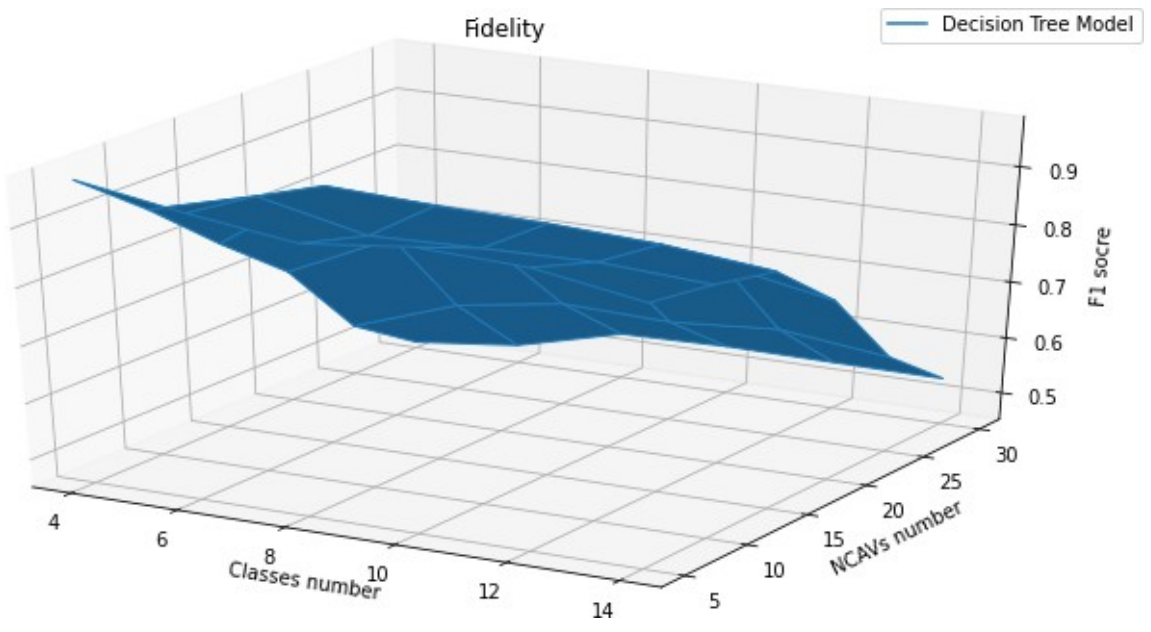

**F I G U R E 10** The fidelity of *DT* model, using*F1 score*

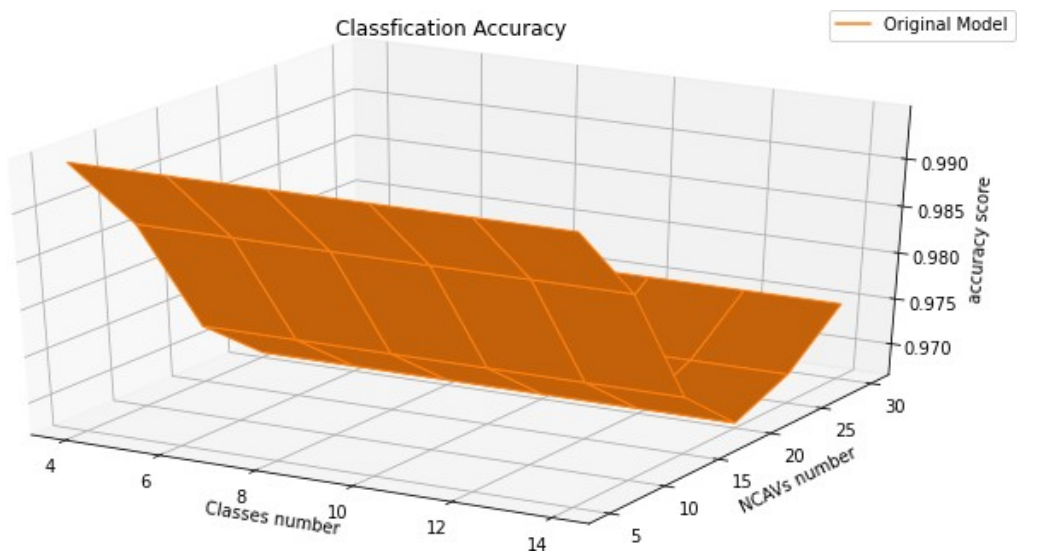

**F I G U R E 1 1** The original model, CNN, classification performance, using *accuracy score*

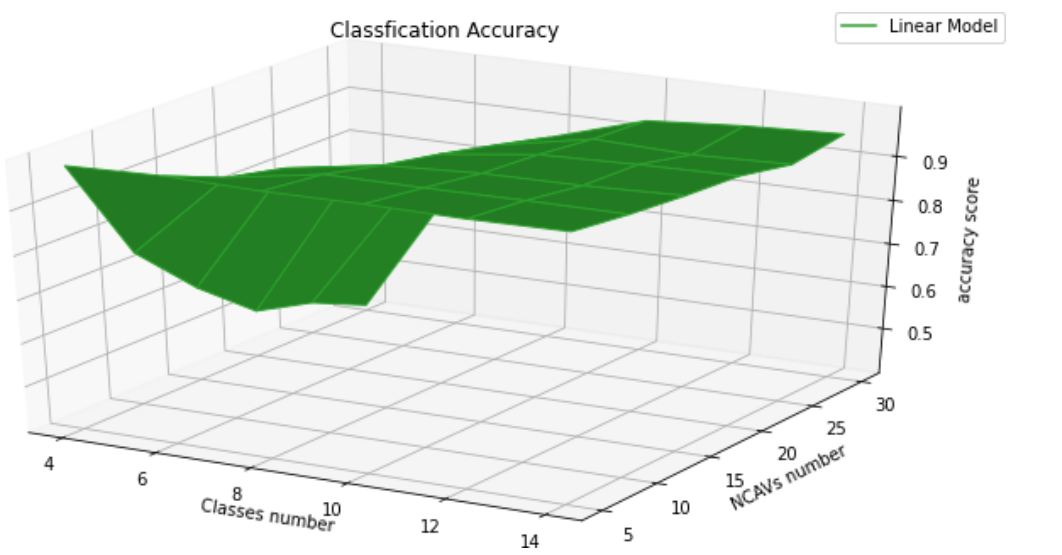

**F I G U R E 1 2** The classification performance of *LR* model, using *accuracy score*

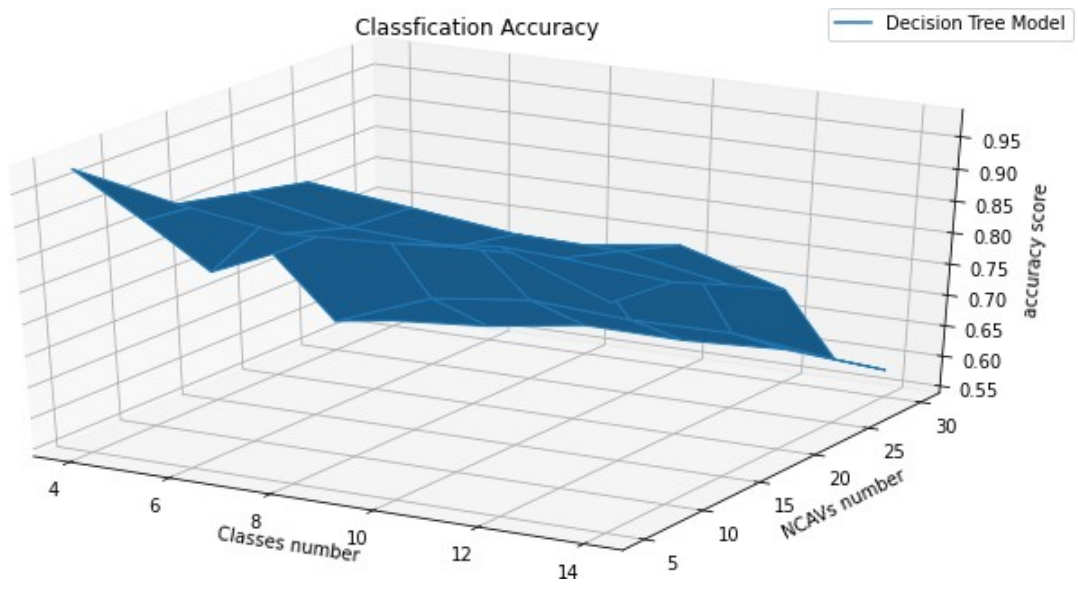

**F I G U R E 13** DT model classification performance, using accuracy score

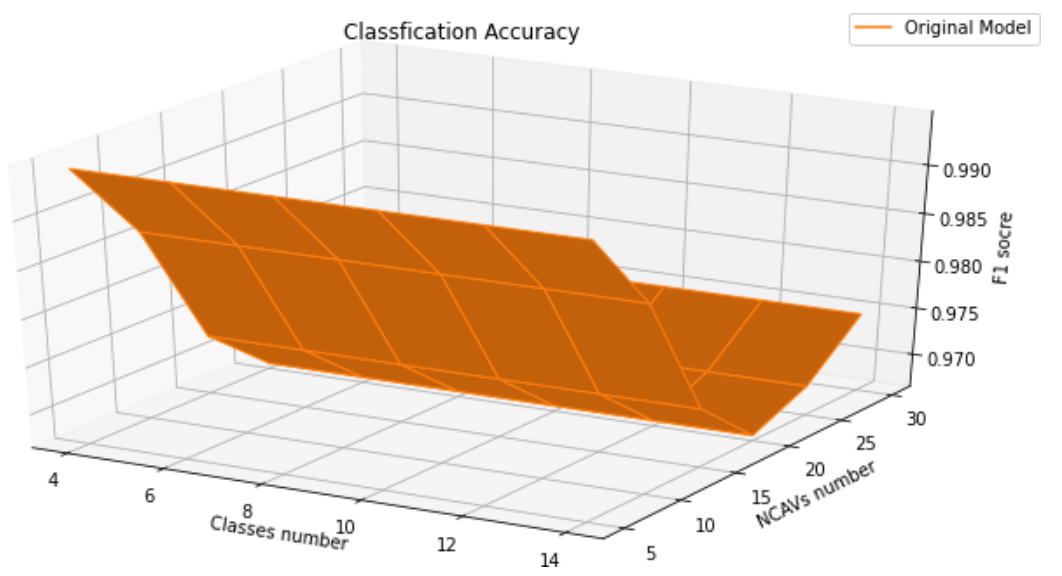

**F I G U R E 14** The original model, CNN, classification performance, using F1 score

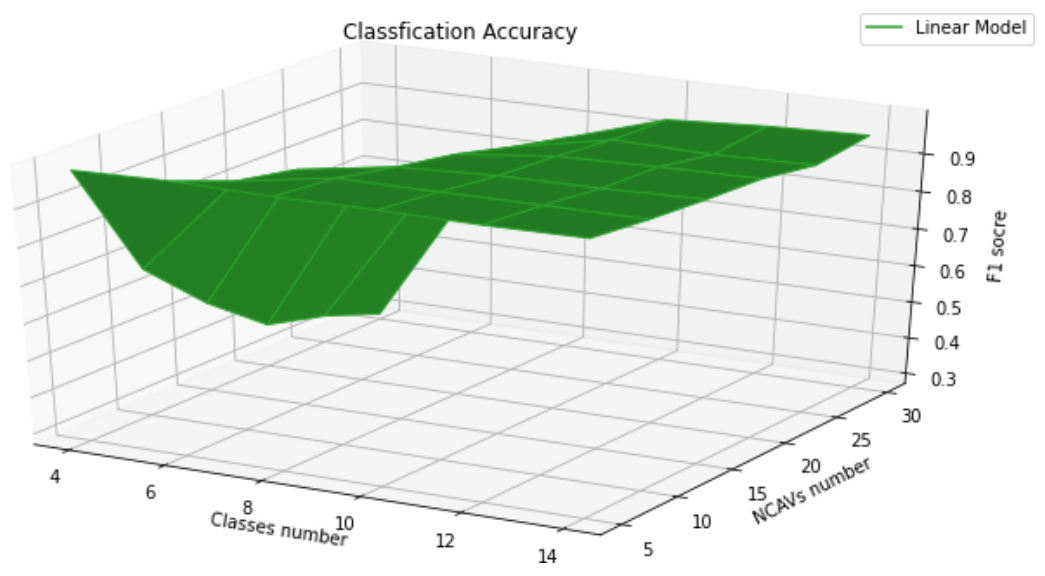

**F I G U R E 15** The classification performanceof the linear model *LR*, using *F1 score*

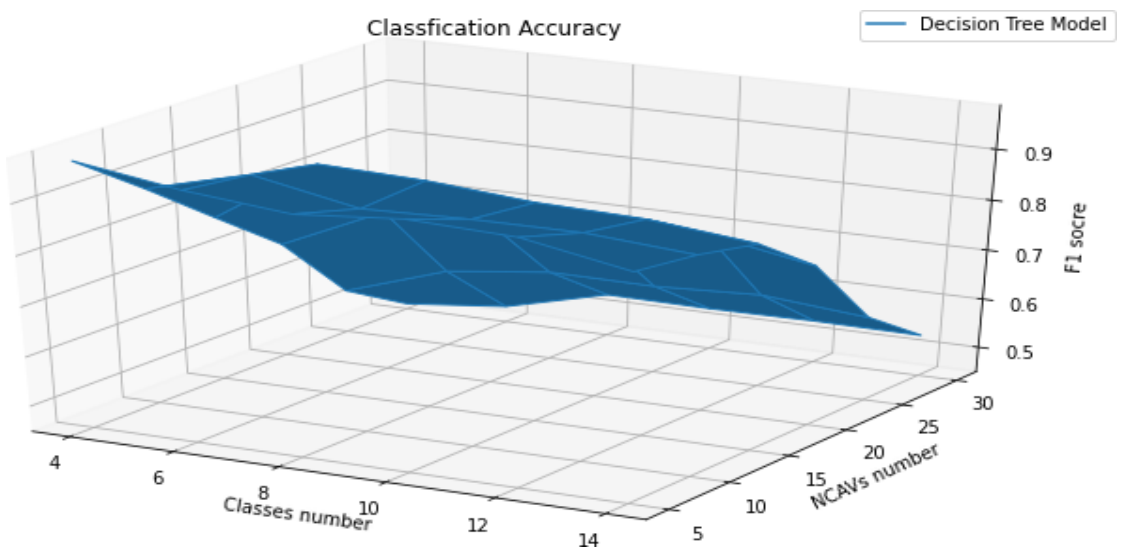

**F I G U R E 16** The decision tree model DT classification performance, using F1 score

From the graphs above we can see that, the results of the two metrics, *accuracy score* and *F1 score*, are remarkably similar for both fidelity and classification performance. We can translate this similarity due to the relatively balanceddistributions across the classes in our target dataset. Further analysis of the graphs indicates that the *F1 score* canhave a slightly lower trend than both models' *accuracy score*.

What is interesting here is that both models have similar patterns. Even though the overall trend of *LR* implies that it could be more faithful and more accurate than the *DT* model, the *LR* trend considerably fluctuates. The fidelity and performance scores of *LR* dramatically drop with $c = 5$. This decrease gets sharper as the parameter $k$ increases. On the contrary, *DT* has a lower overall trend than *LR*, but it does not fall as precipitously as *LR*. The decline points of the *DT* model are much better as they are all above 50%, while for *LR*, the drop reaches 20% and less. Moreover, both models' trends comparatively flatten out after the value $c = 15$. Extracting more than 15 concepts will not improve the model's performance. To have an overall overview, we computed the average fidelity and classification performance over the different values of our two parameters $c$ and $k$ in Table 2. From this table, we can conclude our main computational results as that the *DT* fidelity and performance drops by around 10%, compared to the *LR* explanation model.

| Model | Performance | Fidelity |
|---|---|---|
| Original CNNs Model | 97.75% | N/A |
| *LR* Explanation | 88.50% | 88.93% |
| *DT* Explanation | 78.67% | 78.71% |

**TABLE 2** The average results of the classification performance and fidelity measurements, using *accuracy score*, over the different values of the two used parameters, $c$ and $k$. N/A refers to 'Not Applicable'.

## 4.3 | Human Subject Experiments

Interpretability is an essential requirement to be tested in the context of XAI [19]. A model is described as interpretableand, therefore, trustable when a human can reason about it. That is, a human study is required to further evaluate the performance of explainable models. We conducted human subject experiments using quantitative and qualitative metrics. The ultimate objective is to investigate whether our explanation model (TreeICE) is more interpretable thanthe baseline (ICE) [57].

### 4.3.1 | Experimental Methodology

We start by describing the hypothesis testing procedure we followed.

For our empirical evaluation, we test two main hypotheses:

- **Hypothesis 1:** Decision tree explanation models *DT* are more interpretable than other simple models such as Linear models *LR*. Decision trees can produce more information about the concepts (features) that *M* considers important. The null hypothesis states that both *DT* and *LR* explanation models are equally interpretable by humans : $H_0 : P_L = P_D$. The alternative hypothesis shows that *DT* explanation model is more interpretable andunderstandable than *LR* explanation model: $H_1 : P_D > P_L$. $P$ refers to the proportions of the observed values of correct answers by the participants in task predictions. The goal of the study is to reject the null hypothesis and accept the alternative hypothesis.

- **Hypothesis 2 :** *DT* explanation model provides subjectively better explanations. The null hypothesis is $\boldsymbol{H_0 : P_L = P_D}$, which indicates that *DT* and *LR* explanation models have equivalent quality degree $\boldsymbol{H_0 : P_L = P_D}$. The alternate hypothesis is $\boldsymbol{H_1 : P_D > P_L}$. $\boldsymbol{P}$ denotes the proportion of the observed values of the Likert scale data that refers to the explanation quality data, Explanation satisfaction scale metrics, adopted from Hoffman et al. [23].

To validate the consistency of our experiments and based on the data we obtained, *t* test and Pearson's Chi-squared test were employed, inspiring by similar studies, e.g. [57, 2, 33]. We examined these two different statistical tests due to the variety in the data type and distributions we had.

### 4.3.2 | Human Subject Experiment Design

The dependent variable of the human subject investigation is the interpretability of the explanation model (the inde-pendent variable). As we mentioned earlier, the independent variable has two values, *LR* and *DT* , which are the two experimental conditions in the human study. Interpretability refers to how easily a model and its predictions can be understood by humans [19].

To compare the interpretability of our approach to the baseline, we use *Task prediction* and *Confidence* [23] asquantitative metrics whereas the Explanation Satisfaction Scale [23] represents the qualitative metrics. *Task prediction* is known as an efficient impact on the mental model of humans as it is able to effectively measure the knowledge transformation from an explanation to people [23, 35]. In our experiments, *Task prediction* is defined as that the ability to predict the output decision of the CNN model *M* indicates the understandability of the model decision- making process by a human. While a task prediction may be a cost-effective tool for probing users' mental models,its application should be complemented with a confidence rating, recommended by Hoffman et al. [23]. Therefore, we added the Confidence metric to allow the human subjects to rate their confidence level of the prediction process they have made. In addition, we test the human subjects' satisfaction with the corresponding explanations using the Hoffman et al. [23]'s Explanation Satisfaction Scale in terms of Understanding, Satisfaction, Sufficiency, Completeness, and Trust [23].

In our experiments, Task prediction and Confidence, the quantitative metrics, were performed to investigate hy- pothesis 1 while the Explanation Satisfaction Scale metrics (Understanding, Satisfaction, Sufficiency, Completeness, Trust) were to evaluate hypothesis 2 as the qualitative measurements. We formulate statements of these five metrics in Ta- ble 3, which helps to show the meaning of each metric. t test was used to evaluate Hypothesis 1, so to identify the differences in the quantitative metrics between the two experimental conditions of the two explanation models, LR and DT. Because we established one directional hypothesis, the one-tailed (p value) was chosen for t test. For Hypothesis 2, Pearson's Chi-squared test was employed to investigate the significance results of the Explanation Sat- isfaction Scale measurements [23]. We considered here a non-parametric test not only because of the Likert scale data type but also because the collected data were not normally distributed. The level of the statistical significance is set to ($\alpha = 0.05$) for the purpose of such a study.

We also measure the elapsed time in each experimental condition as an additional quantitative metric since it is one of the important interpretability dimensions [19]. Some domains require making crucial decisions in a short period of time. For instance, specialists suspect an earthquake is imminent. In such cases, the amount of time available to understand an explanation is a necessary aspect to take into consideration and it is also one of the XAI interpretability dimensions [19]. As a result, we recorded the time the subjects spent doing the primary tasks of the experiment that is the 6 tasks in the *testing phase* of the Explanation prediction part. According to Sauro and Lewis [45], the time of usability tasks in related studies considers as a positive skew data due to a possible unusual long time spent on some tasks. Sauro and Lewis [45] suggested that the arithmetic mean becomes a rather poor predictor of the distribution's center whereas the geometric mean provides the most accurate estimate. We calculated both the geometric average time required by the participants to understand the produced explanations in each experimental condition. We also considered reporting the arithmetic mean of the spent time for contrasting purposes.

| | | | | | |
|---|---|---|---|---|---|
| 1. I **understand** how the AI system classifies inputs from this explanation model. | Strongly Disagree<br>1 | Disagree<br>2 | Neither agree nor disagree<br>3 | Agree<br>4 | Strongly Agree<br>5 |
| 2. I am **satisfied** about this explanation model. | Strongly Disagree<br>1 | Disagree<br>2 | Neither agree nor disagree<br>3 | Agree<br>4 | Strongly Agree<br>5 |
| 3. I find the explanation model is **sufficiently** detailed to show how the AI system classifies inputs. | Strongly Disagree<br>1 | Disagree<br>2 | Neither agree nor disagree<br>3 | Agree<br>4 | Strongly Agree<br>5 |
| 4. I find the explanation model is **complete** to show how the AI system classifies inputs . | Strongly Disagree<br>1 | Disagree<br>2 | Neither agree nor disagree<br>3 | Agree<br>4 | Strongly Agree<br>5 |
| 5. I **trust** the AI system because of this explanation. | Strongly Disagree<br>1 | Disagree<br>2 | Neither agree nor disagree<br>3 | Agree<br>4 | Strongly Agree<br>5 |

**TABLE 3** Explanation quality statements we presented to the human subjects. It was set with five point scale in the form of five selected metrics of Hoffman et al. [23]'s Explanation Satisfaction Scale.

### 4.3.3 | Experimental Setup

The experiment was designed in a form of a survey using Qualtrics software. A between subjects design [25] was used to examine the hypotheses we set. Therefore, each participant would be evaluated on only one experimentalcondition. The participants were collected via the Amazon Mechanical Turk (MTurk).

### 4.3.4 | Experimental Tasks

The human subjects were randomly assigned to two groups, for each experimental condition, one for *LR* and the other for *DT* explanation models. Before commencing the survey, the participant was asked to sign a consent form of theethics approval, obtained from the university (The University of Melbourne Human Research Ethics Committee (ID 1749428)). The demographic information was gathered, only age and gender as other private details are not neededin the study. The experiment has two main parts, Part 1 (Explanation prediction), and Part 2 (Explanation quality).

Explanation prediction part has two phases, the *training phase,* and the *testing phase*. We consider the global explanation for the experiment. The human subjects were shown 9 images and 9 explanations in the two phases. The participants were asked to predict the class they think a new input image belongs to, using a corresponding explanation. In other words, the participants were expected to try to understand how to use these explanations in order to predict the images of a certain class before they move to the next question and rate their confidence aboutthe given answer.

Phase 1 of the Explanation Prediction part is the *training phase* which enables the participants to not only under- stand the required task but also learn about the dataset and how to use the explanation model. This phase would assist the participants to and pre-understand and mimic the testing phase, so that they can go through it smoothly. To classify an image to its class label, the participants were expected to pay attention to the features that distinguish a particular class from others. These features can be some concepts, such as eyes, nose, head, .. etc. Therefore, a concept tutorial was first displayed in the *training phase* to highlight examples of these concepts and show how they would look. Then, three training examples, 3 images and 3 explanations, were displayed along with the corresponding expected answer (class) and detailed clarifications of the prediction process. We set the expected answer to be theoriginal model predictions, not the ground-truth labels. This is because we are interested more in testing to what extend the produced explanation mirrors the behavior of the black-box CNN model. Accordingly, it was extremely crucial to ensure the participants understand the target they are evaluating; the model they used is an explanation of another complex AI model. We, as a result, added a related question after the first training example to test their understanding. The correct answer is then provided on the next page to confirm this objective.

Phase 2 is the *Testing phase*, with 6 main tasks and 2 questions for each task. Each task had a unique image and an explanation. The first question involved evaluating the task prediction by asking the participants to predict the classthe given image belongs to, without any help at this stage. This allowed us to test how effectively the training phase and the explanation model had built a mental model of the decision model. The second question involves investigating one of the *Confidence* metric. After answering the first question in each test task, the participants were required to rate their confidence degree. A rating slider of *1* to *5* degrees was set to be answered in ascending order, so *1* refers to the lowest confidence degree and *5* is the highest. Additionally, the participants had the option to view the training examples whenever they wanted while doing the testing tasks. After each test task, the correct answer was displayed in a separate page to show the participants the class the black-box model classifies the image to. However, the participants were not allowed to move back and change their answers.

This would support understanding the black- box model's prediction process; thus, moving to the Explanation Quality part and answering its questions should be more logical. After completing the *testing phase* and answering the six tasks, the Explanation Prediction part is considered as done, and the participants would be ready to move to the Explanation Quality part.

The Explanation Quality part is a group of explanation quality questions to assess the participants' opinion aboutthe explanations in the form of Explanation Satisfaction Scale introduced by Hoffman et al. [23]. The human subjects were asked to rate 5 metrics on a Likert scale of *1* to *5* in order to self-report their feelings about the explanation modelthey used. These quality questions have been positioned in a separate part at the end of the experiment instead of viewing them after each test case. The purpose of that is to ensure that the participants have seen and used differentversions of the assigned explanation model, so better subjective judgment can be made.

The responses of the survey were initially 102 in total, 51 for each experimental condition group. For data quality purposes, only 'master class' workers with 95% or more approval rate were recruited from MTurk platform. The data collected were also cleaned and all the incomplete responses and the ones with less than 10 minutes of the overall time spent were excluded. The average time needed for an experiment is approximately 30 minutes, therefore, responses with much lower time rate indicates that the participant did not pay much attention. The final total number of participants was **70**, 35 for each experimental condition. 64.28% of total participants were males, 34% were females, and 1% preferred to not say their gender. The age of participants in both conditions was between 23 and 63 ($\mu$ = 35.81). We also compensated each participant with US $11 to be automatically received after completing thesurvey.

### 4.3.5 | Experimental Parameters

We randomly drew 9 images from 9 different classes chosen of the dataset, 3 for the training phase and 6 for the testing phase. We decided to have four classes, textitk = 4, for both experimental conditions to balance the complexity and usability of the task. A pre-trial was initially conducted, and it confirmed that less than four classes could be easy topredict the correct class while more than four classes would be quite challenging.

When the number of nodes rises exponentially with the depth of a decision tree model, the outcome of the model becomes less interpretable. The computational analysis we conducted shows that a depth of 10 is a good candidate.That is, a shallower decision tree with fewer branches is favoured for our experimental purposes. The computational results also indicate that both explanation models, *LR* and *DT* , have good performance when the number of extracted concepts *c* is 15 and higher. Therefore, we trained both explanation models with *c* of 15. The explanation model of each task in the Explanation prediction part was trained with different classes. The subjects were also tested with avariety of generated explanations and target images during the study.

## 4.4 | Results and Analysis

In this section, we report the human study findings of both experimental conditions that are the *LR* and *DT* explanation models. We first show the results of the time spent to undertake the tasks on each explanation model. The results of the hypotheses we set are then summarized.

### 4.4.1 | Time Spent of Experiments

Figure 17 contrasts the geometric and the arithmetic average spent time by the participant to complete the *Testingphase*. The participants took a comparable amount of time on both experimental conditions, *LR* and *DT.* It is also worth noting that the participants in the *DT* experiment spent slightly less time doing the test tasks compared to the *LR* explanation. However, there is no significant difference between the two groups, as the *t* test results confirmin Table 5. In other words, *LR* and *DT* explanation models require approximately the same amount of time to be understood.

| | Task Prediction | | | Confidence | | |
|---|---|---|---|---|---|---|
| | *M* | *SD* | *95% CI* | *M* | *SD* | *95% CI* |
| *LR* | 62.8% | 0.23 | [0.54, 0.71] | 3.3 | 0.62 | [3.1, 3.5] |
| *DT* | **74.7%** | 0.24 | [0.66 , 0.83] | **3.9** | 0.65 | [3.6,4.1] |

**TABLE 4** Summary of the descriptive statistical information of the quantitative metrics. M: Mean, SD: Standard deviation, 95% CI: 95% Confidence interval.

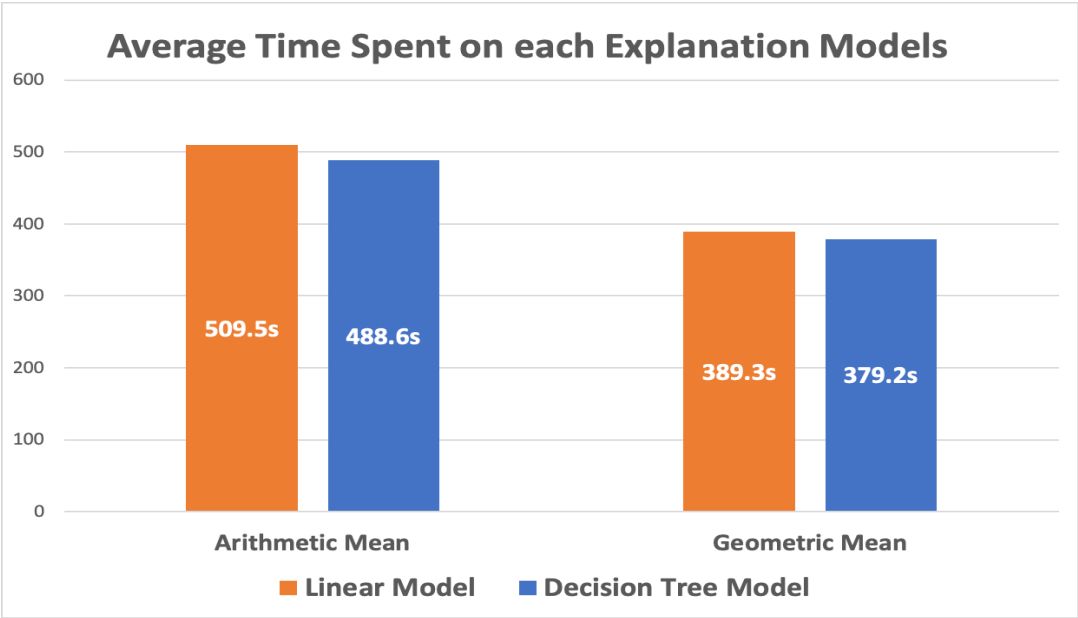

**FIGURE 17** The computed arithmetic and geometric means of the total time required to complete the testing phase in both experimental conditions. Y-axis represents the average time in seconds. Less is better.

### 4.4.2 | Hypothesis 1

For hypothesis 1, we aim to examine whether *DT* explanation model leads to a better understanding of the black-boxCNN model. As mentioned earlier, *Task prediction* and *Confidence* metrics were used to investigate hypothesis 1. For *Task prediction*, we refer to the accuracy of the participants' correct responses in question 1 of each test task in the *Testing phase*. For the *Confidence*, question 2 of each test task was used to collect the data of this metric.

The key descriptive statistics of the data for hypothesis 1 metrics are gathered in Table 4. It can be noted that the results of the mean, standard deviation and the 95% confidence interval are all in favor of the *DT* explanation model.

The participants who undertook the explanation of the *DT* model had a substantially better average of correct predictions. Moreover, *t* test results of *Task prediction* verifies that the *DT* model has statistically significant results atthe 0.05 level ((*p* value) = 0.02), as shown in Table 5. As a result, we can deduce that the *DT* model outperforms the *LR* in the *Task prediction*. To convey the uncertainty [29] and to substantiate the statistical significance result we obtained for the *task prediction* metric, the 95% confidence interval (*95% CI*) for the mean of correct answers was also computed as the error bars indicate in Figure 18. Since the ranges of the (*95% CI*) in both conditions are relatively not wide and they do not contain 0, we can certainly confirm the significance outcome.

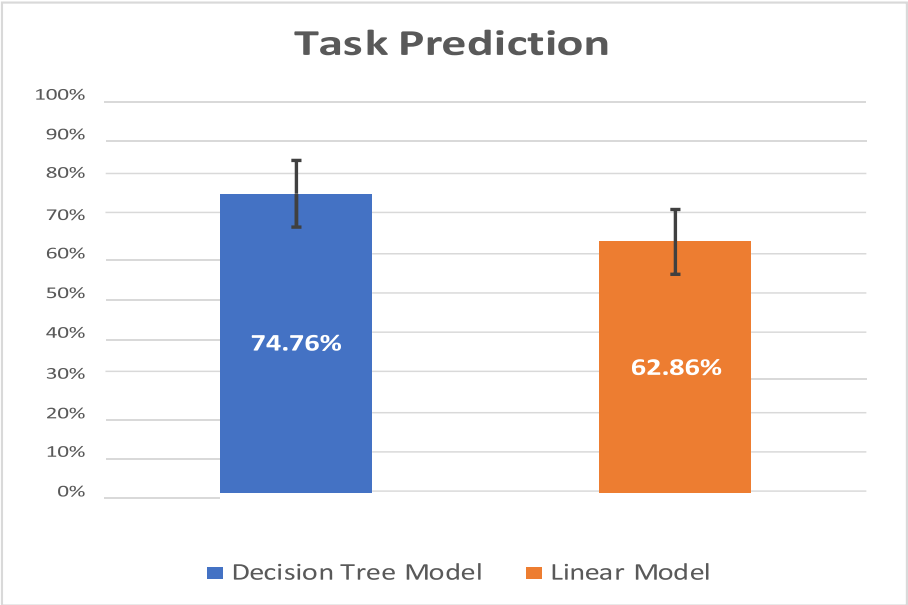

**FIGURE 18** The average of the correct responses by the participants on the two experimental conditions. The decision tree experiment records more accurate predictions by the participants across the test tasks than the linearmodel experiment. Error bars represent the (95% CI).

The results of question 2 mirror the outcome we obtained from measuring *Task prediction*. That is, the confidence degree for those who rate their confidence with the highest degree, 5, was **24.7%** for the *DT* explanation model whileit drops to 17.6% for the *LR* explanation model. Interestingly, 71.9% of the subjects rate their confidence with 4 and 5 degrees in the Decision Tree model compared to 51.4% in the Linear model. Figure 19 reveals the confidence rates distribution for each test task in both explanation models. Furthermore, it can be evident, from Table 5, the (*p* value)of the *t* test supports these results as the difference is highly significant for the *Confidence* metric.

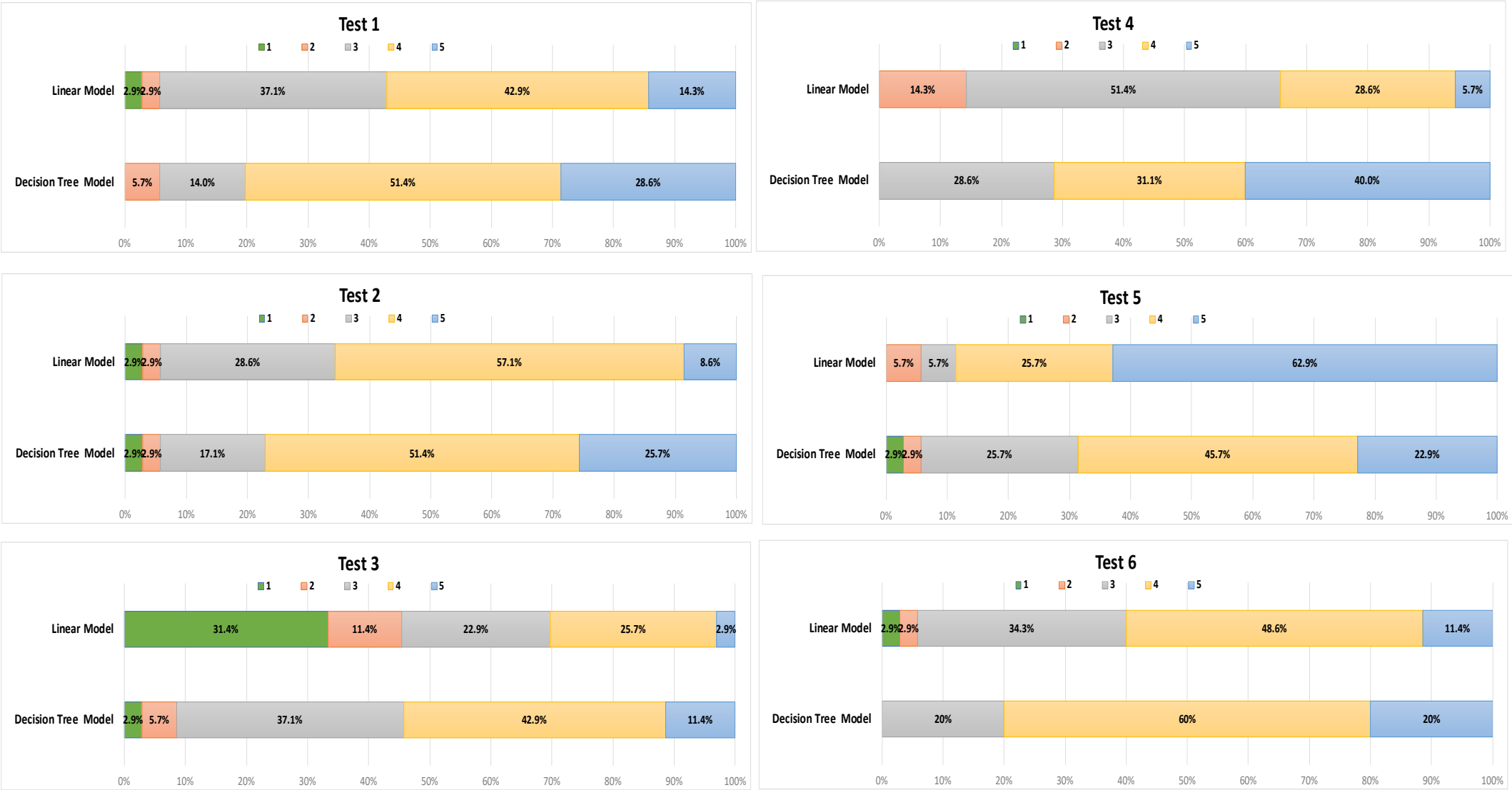

**FIGURE 19** The data counts of the confidence rate reported for the six test cases of the *testing phase* across thetwo explanation models.

Given the above findings, we accept $H_1$and reject $H_0$ for hypothesis 1. The results of both *Task prediction* and *Confidence* metrics verify that *DT* performs better than *LR,* so *DT* explanation model is more interpretable.

| Metric | *t* stat | *p* value |
|---|---|---|
| Task Prediction | -2.04 | **0.02** |
| Confidence | 3.74 | **0.0001** |
| Time Spent | -0.28 | 0.39 |

**TA B L E 5** *t* test shows significance result at level 0.05 for *Task prediction* and *Confidence* metrics, which conclude that *DT* model provides more understandable explanations.

### 4.4.3 | Hypothesis 2

Hypothesis 2 was set to test the quality of the two explanation models we compare, *DT* and *LR.* The Likert scale data of the five metrics were collected from the Explanation quality part of the two experimental conditions and their distributions are visualized in Figure 20. What stands out in this figure is that most of the subjects in the *DT* experimental condition rated the five explanation quality metrics with high scores. We can also note that the mean of each metric is higher in *DT* explanation model than in the *LR,* as Table 6 demonstrates. Hence, it is obvious that *DT* explanation model has a positive trend across the five Satisfaction Explanation Scale metrics. However, we cannot completely accept $H_1$ as not all the metrics results are significant. This was further investigated using the Chi-squared test, as Table 7 shows. While the Chi-square test did not show any significant differences between *DT* and *LR* in terms of *Understandability*, *Completeness*, and *Trust* metrics, the *Satisfying* and *Sufficiency* have statically significant results. That is, the *DT* model provides more satisfactory and detailed explanations than the *LR* model.

| | **Satisfaction Explanation Scale** | | | | | | | | | | | | | | |
|---|---|---|---|---|---|---|---|---|---|---|---|---|---|---|---|
| | Understandability | | | Satisfying | | | Sufficiency | | | Completeness | | | Trust | | |
| | *M* | *SD* | *95% CI* | *M* | *SD* | *95% CI* | *M* | *SD* | *95% CI* | *M* | *SD* | *95% CI* | *M* | *SD* | *95% CI* |
| *LR* | 4.2 | 0.75 | [3.9, 4.4] | 4.2 | 0.74 | [4, 4.5] | 4 | 0.85 | [3.7, 4.3] | 3.8 | 0.96 | [3.4, 4.1] | 4.1 | 0.90 | [3.8, 4.4] |
| *DT* | **4.5** | 0.81 | [4.2, 4.8] | **4.5** | 0.85 | [4.2, 4.8] | **4.4** | 0.88 | [4.1, 4.7] | **4.2** | 0.88 | [3.9, 4.5] | **4.4** | 0.73 | [4.1, 4.7] |

**TA B L E 6** Summary of the descriptive statistical information of the qualitative metrics. M: Mean, SD: Standard deviation, 95% CI: Confidence interval.

| Quality Metric | Pearson's Chi-squared | |
|---|---|---|
| | $X^2$ | *p* value |
| Understandability | 6.065 | 0.11 |
| Satisfying | 9.441 | **0.02** |
| Sufficiency | 11.252 | **0.02** |
| Completeness | 8.775 | 0.06 |
| Trust | 2.090 | 0.55 |

**TABLE 7** Pearson's Chi-squared Test for the five explanation quality metrics of the Explanation Satisfaction Scaleby [23]

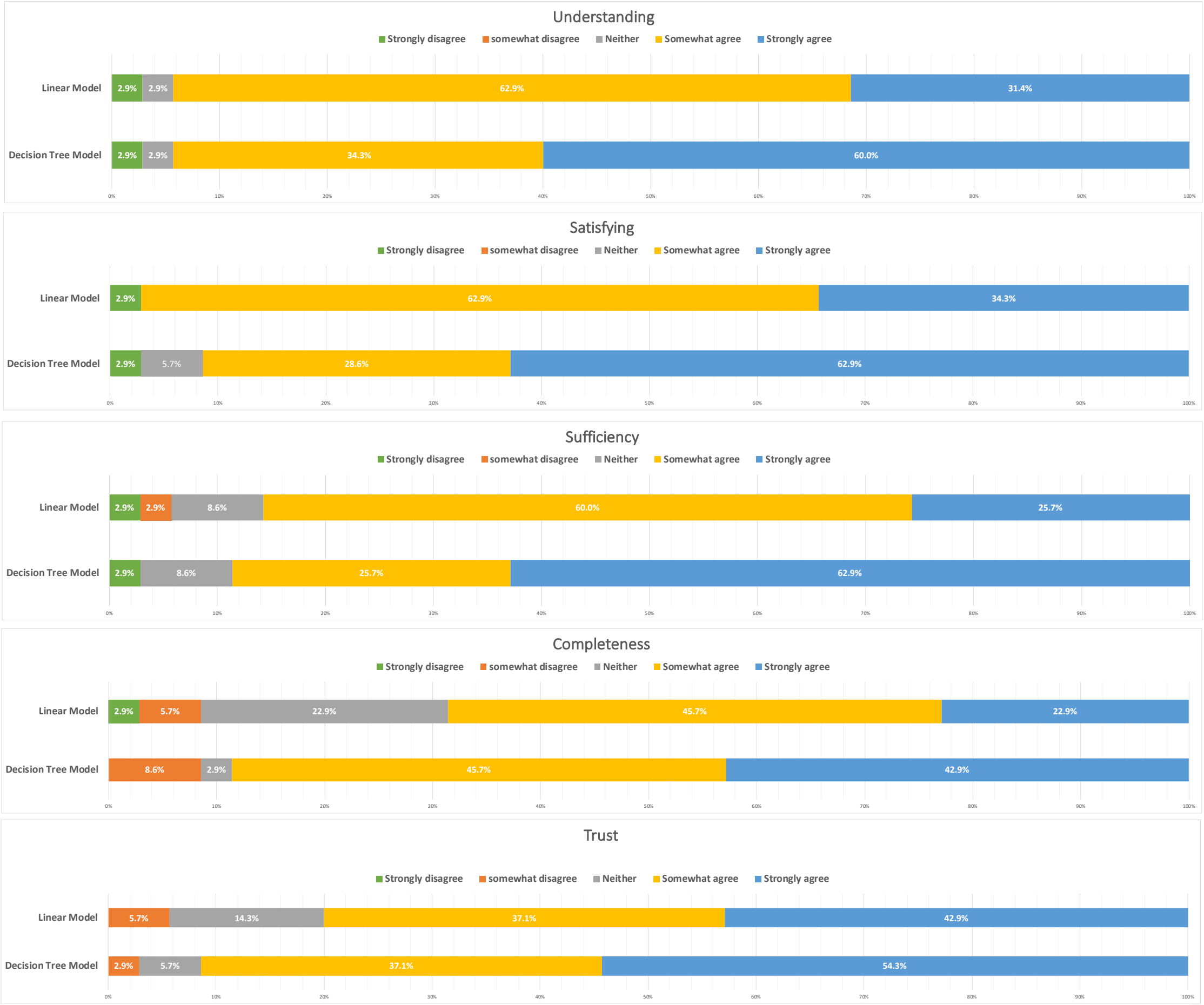

**FIGURE 20** Frequencies data of the five qualitative quality metrics across the *DT* and *LR* explanation models.

# 5 | DISCUSSION

From the human subject experiment results in Section 4.4, we deduce that the decision tree explanation is more interpretable than the linear explanations. The quality of the decision tree explanation is at least equivalent to the linear explanation's quality. We referred to the explanation quality to the five measures we used from the Explanation Satisfaction Scale by Hoffman et al. [23]. Those are Understanding, Satisfaction, Sufficiency, Completeness, and Trust. Even though the Likert scale counts of all the five metrics were in favor of the decision tree model, the statistical test did not reveal significant results on all of them. Nevertheless, the decision tree model shows significantly better results than the linear model in terms of the satisfaction degree and the sufficient information provided. It can therefore be assumed that the participants feel that the decision tree explanation is more satisfactory and has more sufficient details than the linear model explanations. These findings support other studies that hired a decision tree to interpret the decision-making process of CNN models, e.g. [56, 37, 24].

From the computational assessment, the presented results in Section 4.2.1 show that there has been a declinein both fidelity and classification performance of the decision tree explanation model with around 10% lower than the linear model. Interestingly, both explanation models record the highest classification performance and fidelity scores with 4 trained classes or less ($k \leq 4$), regardless of the number of the extracted concepts ($c$). However, $c$ withvalue of 5 performs the worse as number of classes ($k$) grows. Therefore, it can be inferred that for both linear and decision tree approximations, more than 10 reduced channels or extracted concepts ($c \geq 10$) is required to optimizethe explanation algorithm with any number of classes. It was also noted that the *DT* explanation performs better than *LR* with small $c$. In other words, when the number of extracted concepts is inadequate, a decision tree explanation model would be more effective.

One unanticipated finding was the average time required to complete the test tasks of each explanation modelwas almost alike. The presentation form and the visualization nature of the decision tree and the linear model greatlydiffer. In the linear experiment, users were expected to step through each class to analyze its concepts and classify the provided test image. In contrast, tree-like structures cope with this shortcoming as participants can view the extracted concepts in one form. Because of that, we expected that users of the tree explanation would take less time. However, the difference in the average spent time by the subjects for the two experimental conditions was negligible, as Figure 17 demonstrates. We believe the number of target classes, we used for the human experiments, caused the similarity in the spent time. If there were more classes, above 4, linear models could have been more complicated to use.

Considering the duration of the explanation models experiments opens the door to another important aspect ofinterpretable models. It is the explanation's complexity or size of a model, according to Guidotti et al. [19]. Even though complex models can be more accurate, the interpretability degree of such models would decrease. Thatleads us to the so-called trade-off between interpretability and performance. Indeed, that was the main reason thatdrove us to not analyze the performance of our proposed framework before investigating the decision tree classifiersize. By experimenting with several maximum depths of the tree, we ended up with a shallow decision tree that is meaningful and usable. We were confidently able to make this decision as deeper trees did not enhance our model in terms of either understandability or accuracy, as seen in Figure 6.

Even though the balance of the interpretability and performance trade-off may be challenging to attain, our present results are significant in several respects. Decision tree-based explanations are still valuable, especially when interpretability is more relevant than the system's accuracy. Most importantly, the current focus in related industries is to have AI models that are transparent and interpretable rather than just extraordinary accurate models. In many cases, a slight drop in a model's accuracy would not dramatically impact the application's overall performance. Webelieve that a powerful AI system today is the system that enables decision-makers to understand, use, and trust them.

In addition, most of the former related works, which explain CNN models using decision trees, show the nodes of their decision tree model based on the extracted features rules in a more specialized manner. They do not often visualize the related images that contain an underlying feature in the nodes. However, TreeICE visualizes the features in a meaningful tree-like structure; each node consists of a conceptual feature that is visibly highlighted in corresponding image samples so that non-expert users should be able to use and understand.

To illustrate the extracted concepts (NCAVs) generated based on a decision tree and how they differ from the original linear model, we present an example of a local explanation using both models to explain the input image shown in Figure 21. In Figure 22, an explanation example generated by the decision tree from our framework TreeICE is illustrated while Figure 23 depicts an explanation derived by the linear approximation of ICE method. Four classesare trained to generate these explanations for both models. The extracted concepts of each trainedclass generated from ICE are displayed separately in a linear way along with the corresponding concept importanceweight. In contrast, the decision tree generates an explanation in a human-readable tree structure.

## 5.1 | Limitations and Future Work

Despite the evidence of this work capability in contributing to the stated research problem, some shortcomings should be highlighted. A variety of potential directions for future extensions and investigations are worth considering. For instance, symbolic reasoning over a decision tree can be used to derive counterfactual explanations, inspired by [18]. Counterfactual algorithms provide an explanation by creating a contrastive scenario for the human to understand the underlying process through cognitive demonstration of the difference [53]. Therefore, having these explanations extracted from our generated decision tree model would strengthen the explainability of our framework. Although we used a professional benchmark dataset, applying the TreeICE algorithm to more critical data like medical imaging will show the work in a broader context of a real-world application. We employed only one of the CNN well-known architectures; however, other CNN architectures can highly be relevant to generalize our findings. Our experiments show that there is still a scope of improvement in the performance of the decision tree model so that we could have more accurate and faithful explanations. We have compared our proposed decision tree approximation to the corresponding approximation used by the original work, ICE [57]. It will be interesting to develop a comparative study that compares our framework with other recent concept-based explanation algorithms in the field. Furthermore, we would like to investigate some theoretical characteristics, such as the optimal sample size and computational improvements, such as parallelization and GPU processing. It would help to facilitate the required real-time explanations efficiently.Lastly, to deliver a fully functional XAI system, an interface of the explainable model should be deemed necessary forend-users in real-world applications.

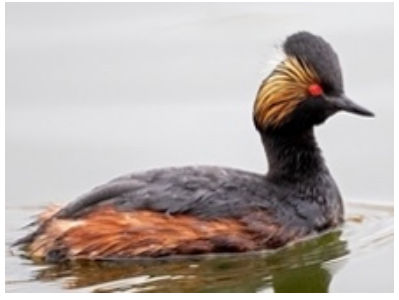

**F I G U R E 21** An example of an input image used to generate a concept-based explanation by the two comparative algorithms, ICE and TreeICE.

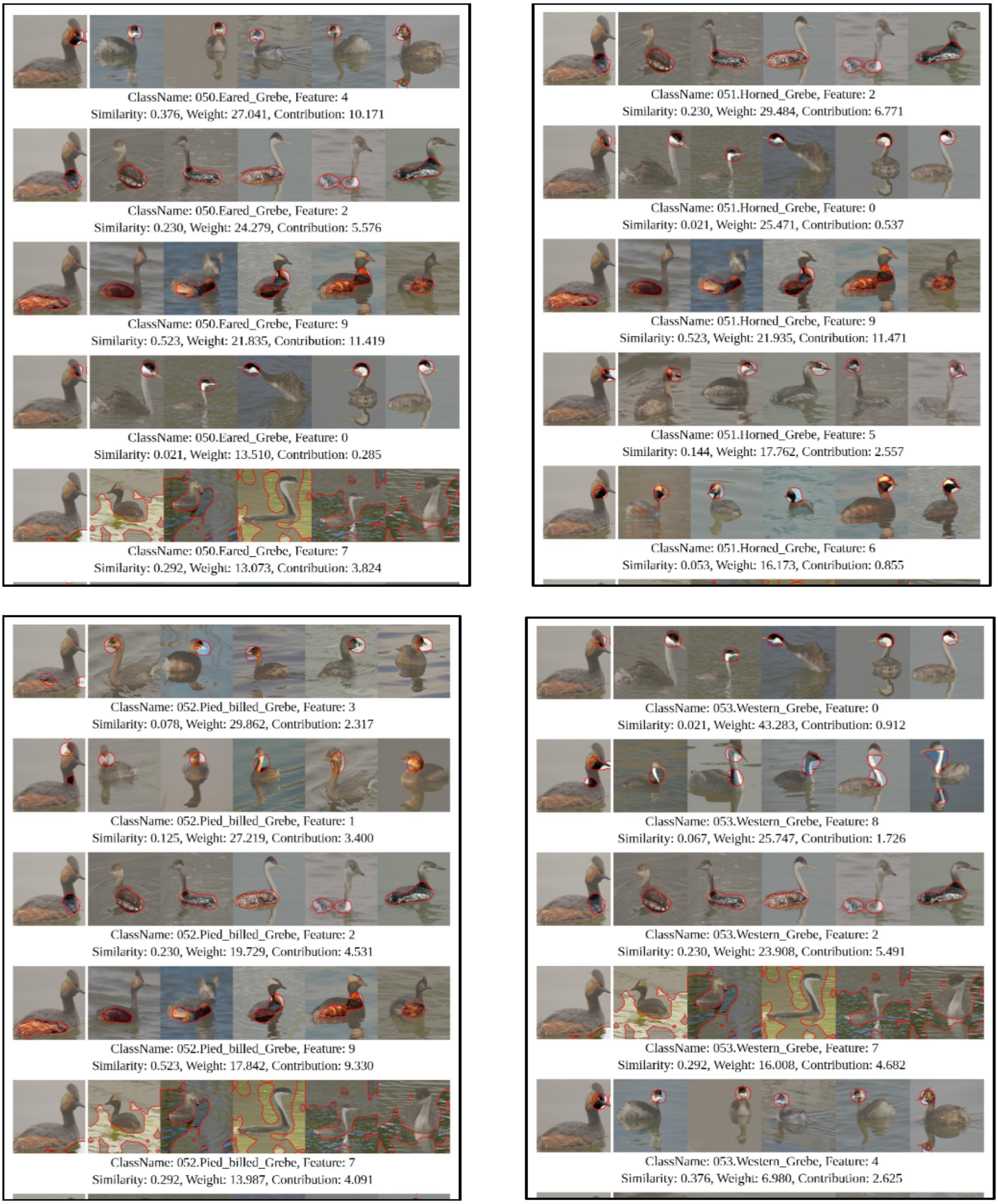

**F I G U R E 2 2** A linear explanation generated by ICE algorithm Zhang et al. [57] to explain the prediction of the input image, shown in Figure 21. Similarity score refers to the degree of similarity between the concept extracted from the input image and the concept of the class-specific shown in 5 samples. Weight means the concept's importance to the class.

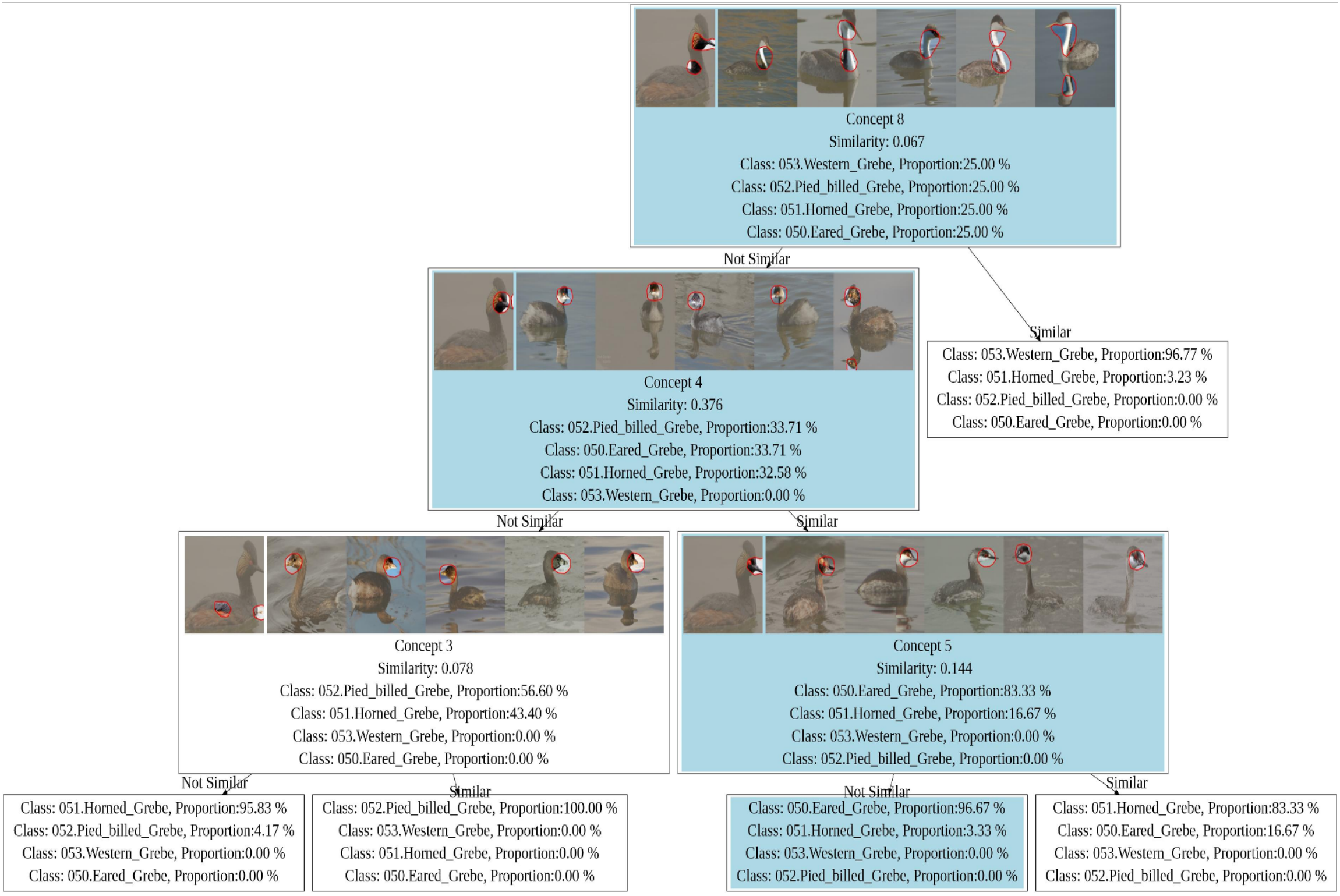

**FIGURE 23** A decision tree explanation generated by our algorithm, TreeICE, to explain the prediction of the input image, displayed in Figure 21. Similarity score refers to the degree of similarity between the concept extractedfrom the input image and the concept of the class-specific shown in 5 samples. Proportion indicates the probabilitythat a derived concept belongs to the corresponding class.

## 6 | CONCLUSION

This work investigated whether one of the state-of-the-art concept-based explanation algorithms can be combined with a decision tree model instead to improve fidelity, performance, and interpretability. We used ICE as a concept- based explainer. ICE generates high-level features (extracted concepts) from the non-negative factorization matrix (NMF) reducer and uses a linear model to approximate a target model. We targeted replacing their linear approximation with a decision tree model. Our framework, TreeICE, aimed to facilitate the use of the extracted concepts to reform the concept-based explanation algorithm (ICE) and effectively interpret a CNN model. We hypothesized that the decision tree would produce a more natural explanation that should be more understandable and interpretable. We proposed a coherent and usable concept-based explanation framework, TreeICE, that generates automaticallyglobal and local explanations with quantitative and qualitative evaluation metrics. To the best of our knowledge, no prior work in the field produces the most recent concept-based explanations, ICE, and integrates it with a more natural and understandable algorithm. Unlike similar works in this area, TreeICE visualizes its final explanations in a meaningful tree-like structure so that human queries can be smoothly answered, and complex systems can be trusted accordingly.

## Acknowledgements

This article has benefited from Dr. Ruihan Zhang (The University of Melbourne) who provided useful discussions onthe ICE framework we extended in this work.

## Conflict of interest

The authors declare no conflict of interest.